\documentclass[10pt,twocolumn,letterpaper]{article}

\usepackage[pagenumbers]{cvpr} 
\usepackage{chapterbib}
\usepackage{algorithm}
\usepackage[noend]{algpseudocode}
\usepackage{setspace}
\usepackage{etoolbox}
\usepackage{multirow}
\usepackage{makecell}
\usepackage{booktabs} 
\usepackage{comment}
\usepackage{colortbl}
\usepackage{array}  
\usepackage{graphicx}  
\usepackage{amsmath} 
\usepackage{amsthm}
\usepackage{tabu}
\usepackage{adjustbox}
\usepackage{arydshln} 
\usepackage{caption}
\usepackage{wrapfig} 
\newcommand{\confhdr}[1]{\makebox[1em]{\textsc{#1}}}
\newcommand{\low}[1]{\raisebox{0pt}[0pt][0pt]{#1}}
\newcommand{\s}{\hphantom{0}}

\newcommand{\redulbf}[1]{\underline{\textbf{\textcolor[rgb]{1,0,0}{#1}}}}
\newcommand{\bluebf}[1]{\textbf{\textcolor[rgb]{0,0,1}{#1}}}
\newcommand{\ud}{\mathrm{d}}
\newcommand{\bfx}{\boldsymbol{x}}
\newcommand{\bff}{\boldsymbol{f}}
\newcommand{\bfI}{\boldsymbol{I}}
\newcommand{\bfQ}{\boldsymbol{Q}}
\newcommand{\bfK}{\boldsymbol{K}}
\newcommand{\bfV}{\boldsymbol{V}}
\newcommand{\bfX}{\boldsymbol{X}}
\newcommand{\bfW}{\boldsymbol{W}}
\newcommand{\bfn}{\boldsymbol{n}}
\newcommand{\bfd}{\boldsymbol{d}}
\newcommand{\bfm}{\boldsymbol{m}}
\newcommand{\bfeps}{\boldsymbol{\epsilon}}
\newcommand{\bfmu}{\boldsymbol{\mu}}
\newcommand{\bfomega}{\boldsymbol{\omega}}
\newcommand{\bfSigma}{\boldsymbol{\Sigma}}
\newcommand{\bbR}{\mathbb{R}}
\newcommand{\bbE}{\mathbb{E}}
\newcommand{\matN}{\mathcal{N}}
\newcommand{\tilp}{\tilde{p}}
\newcommand{\bftilx}{\boldsymbol{\tilde{x}}}

\newcommand{\bleftBrackets}{\big(}
\newcommand{\brightBrackets}{\big)}
\newcommand{\leftBrackets}{\left(}
\newcommand{\rightBrackets}{\right)}
\newcommand{\leftSquareBrackets}{\left[}
\newcommand{\rightSquareBrackets}{\right]}
\newcommand{\BBleftSquareBrackets}{\Bigg[}
\newcommand{\BBrightSquareBrackets}{\Bigg]}
\newcommand{\tmu}{\text{mu}}
\newcommand{\tdata}{\text{data}}
\newcommand{\tcov}{\text{cov}}
\newcommand{\cin}{c_{\text{in}}}
\newcommand{\cout}{c_{\text{out}}}
\newcommand{\cskip}{c_{\text{skip}}}
\newcommand{\cnoise}{c_{\text{noise}}}
\newcommand{\hatt}{\hat{t}}
\newcommand{\hatbfx}{\hat{\bfx}}
\newcommand{\Schurn}{S_\text{churn}}
\newcommand{\Stmin}{S_\text{tmin}}
\newcommand{\Stmax}{S_\text{tmax}}
\newcommand{\Snoise}{S_\text{noise}}
\newcommand{\lB}{\left\{}
\newcommand{\rB}{\right\}}
\newcommand{\lb}{\leftBrackets}
\newcommand{\rb}{\rightBrackets}
\newcommand{\lsb}{\leftSquareBrackets}
\newcommand{\rsb}{\rightSquareBrackets}

\newcommand{\lL}{\left\|}
\newcommand{\rL}{\right\|}
\newcommand{\termx}{\lb\bfx(0) + \cfrac{1-s}{s}\bfmu^l+\bfn^l\rb}
\newcommand{\cskips}{\cskip\lb\sigma\rb}
\newcommand{\cins}{\cin\lb\sigma\rb}
\newcommand{\couts}{\cout\lb\sigma\rb}
\newcommand{\cnoises}{\cnoise\lb\sigma\rb}
\newcommand{\lambdas}{\lambda\lb\sigma\rb}

\newcommand{\mL}{\mathcal{L}}
\newcommand{\Ftheta}{F_{\theta}}
\newcommand{\meantermx}{\text{mean}\lb\lB\cskips \termx\rB_{l=1}^L\rb}
\newcommand{\Fthetax}{F_{\theta}\lb \lB\cins\bftilx^l(t)\rB_{l=1}^L;\cnoises;c\rb}

\newcommand{\bbEsxn}{\bbE_{\sigma,\bftilx_0(t),\bfn}}
\newcommand{\sigmatmax}{\sigma_{\text{max}} }
\newcommand{\sigmatmin}{\sigma_{\text{min}} }

\newcommand{\tvar}{\text{Var}}
\newcommand{\termxwobn}{\bfx(0) + \cfrac{1-s}{s}\bfmu^l}
\newcommand{\termxwob}{\bfx(0) + \cfrac{1-s}{s}\bfmu^l + \bfn^l}
\newcommand{\termxwolbn}{\bfx(0) + \cfrac{1-s(t)}{s(t)}\bfmu}
\newcommand{\tvarxmun}{\tvar_{\bfx(0),\bfmu^l,\bfn^l}}
\newcommand{\tvarn}{\tvar_{\bfn^l}}
\newcommand{\tvarx}{\tvar_{\bfx(0)}}
\newcommand{\tvarmu}{\tvar_{\bfmu^l}}
\newcommand{\tvarxmu}{\tvar_{\bfx(0),\bfmu^l}}
\newcommand{\tcovar}{\text{Cov}}
\newcommand{\termmu}{\cfrac{1-s}{s}\bfmu^l}
\newcommand{\sumxmun}{\cfrac{\cskips}{L}\sum_{l=1}^{L}{\termx}}
\newcommand{\termfracs}{\lb\cfrac{1-s}{s}\rb}
\newcommand{\fracs}{\cfrac{1-s}{s}}
\newcommand{\st}{s(t)}
\newcommand{\sigmat}{\sigma(t)}
\newcommand{\sigmatd}{\sigma_{\text{data}}}
\newcommand{\sigmatmu}{\sigma_{\text{mu}}}
\newcommand{\sigmacov}{\sigma_{\text{cov}}}
\newcommand{\blb}{\bleftBrackets}
\newcommand{\brb}{\brightBrackets}
\newcommand{\icksips}{{\blb 1 - \cskips \brb}}
\newcommand{\cskipsL}{\cfrac{\cskips}{L}}
\newcommand{\summu}{\sum_{l=1}^{L}{\bfmu^l}}
\newcommand{\argmin}{{\arg\min}}

\newcommand{\perturbkernel}{
p_{0t} \blb\bfx(t)\mid \bfx(0),\bfmu\brb
}
\newcommand{\udxi}{{\ud\xi}}
\newcommand{\bftxzt}{\bftilx_0(t)}
\newcommand{\dotst}{\dot{s}(t)}
\newcommand{\dotsigmat}{\dot{\sigma}(t)}
\newcommand{\nx}{\nabla_{\bfx}}
\newcommand{\score}{\nx \log p_{t}(\bfx)}
\newcommand{\udt}{\ud t}
\newcommand{\udbx}{\ud \bfx}
\newcommand{\bftxt}{\bftilx(t)}
\newcommand{\Dt}{D_{\theta}}
\newcommand{\ntx}{\nabla_{\bftxt}}
\newcommand{\tscore}{\ntx \log p_t \blb\bftxt\brb}
\newcommand{\kt}{k(t)}
\newcommand{\cs}{\hspace{0.5mm}}%

\newcommand{\hatti}{\hatt_i}
\newcommand{\hatxil}{\hatbfx_i^l}
\newcommand{\DF}{D_F}

\newcommand{\pt}{p_{t}}
\newcommand{\scoreq}{\nabla_{\bftxt}\log \qbftxt}
\newcommand{\Eq}{\bbE_{q_t\lb\bftxt\rb}}
\newcommand{\Eqtxzt}{\bbE_{q_t\lb\bftxzt\rb}}
\newcommand{\Eqtxtc}{\bbE_{q_t\lb\bftxt\mid \bftxzt\rb}}
\newcommand{\fisher}{\DF \lb q_t\blb\bftxt\brb \parallel \pt\blb\bftxt \brb \rb}
\newcommand{\qcbftxt}{q_t\blb\bftxt\mid \bftxzt\brb}
\newcommand{\scoreqc}{\nabla_{\bftxt}\log \qcbftxt}
\newcommand{\qbftxt}{q_t\blb\bftxt\brb}
\newcommand{\qbftxzt}{q_t\blb\bftxzt\brb}
\newcommand{\pbftxt}{p_t \blb\bftxt\brb}
\newcommand{\const}{\text{const}}
\newcommand{\nbftxt}{\nabla_{\bftxt}}
\newcommand{\tpid}{{\lb2\pi\rb}^{-\frac{d}{2}}}
\newcommand{\dcovar}{\det \lb\sigmat^2\bfI\rb}
\newcommand{\txtstxzt}{\blb\bftxt - \bftxzt\brb}
\newcommand{\Npower}{\lb
            -\frac{1}{2} 
                {\txtstxzt}^{T} {\blb\sigmat^2\bfI\brb}^{-1}
                {\txtstxzt}
        \rb}
\newcommand{\Eqxtxz}{\bbE_{q_t\lb\bftxt,\bftxzt\rb}}
\newcommand{\Dtfunc}{\Dt\blb\bftxt;\sigmat;c\brb}
\newcommand{\Dtfuncn}{\Dt\blb\bftxzt+\bfn;\sigmat;c\brb}
\newcommand{\jointq}{q_t\lb\bftxt,\bftxzt\rb}
\newcommand{\pdata}{p_{\text{data}}}
\newcommand{\Lt}{L\blb \Dt,\sigmat\brb}
\newcommand{\allimg}{{\lB \hat\bfx_{i}^l\rB}_{l=1}^L}
\newcommand{\allimgN}{{\lB \hat\bfx_{N}^l\rB}_{l=1}^L}
\newcommand{\PSNR}{\text{PSNR}}
\newcommand{\SSIM}{\text{SSIM}}
\newcommand{\SAM}{\text{SAM}}
\newcommand{\MAE}{\text{MAE}}
\newcommand{\LPIPS}{\text{LPIPS}}
\newcommand{\RMSE}{\text{RMSE}}
\newcommand{\bfy}{\boldsymbol{y}}
\newcommand{\bfhaty}{\hat \bfy}

\newcommand{\mF}{\mathcal{F}}

\newcommand{\network}{
\begin{figure*}[t]
  \centering
    \includegraphics[width=1.0\linewidth]{figs/network.pdf}
    \caption{
    Illustration of the denoising network. (a) The network concurrently denoises sequences of noisy cloudy images (\textit{noisy mean}), cloudy images (\textit{mean}), and optional auxiliary modal images (\textit{aux}) to generate results (\textit{pred}). 
    The notation $\times L$ indicates $L$ weight-sharing copies. 
    (b) We extend the original HDiT Blocks 
    to THDiT Blocks to integrate temporal information. 
    (c) TFSA collapses the temporal dimension of inputs and generates the attention masks. For simplicity, we present a single-head scenario. Feature map dimensions are indicated below each block, where $N$ is the batch size, $H$ is the height, $W$ is the width, $L$ is the sequence length, $C$ is the channels of feature maps, $G$ is the number of heads, $d_c$ is the channels of condition vectors, and $d_k$ is the channels of query and key matrices.
    }
    \label{fig:network}
    \vspace{-2mm}
\end{figure*}
}
\newcommand{\trainCode}{
\begin{algorithm}[t]
    \footnotesize
    \caption{Our training step with $s(t) = {1}/{\lb1+\alpha t\rb}$ and $\sigma(t)=t$.}
    \label{alg:train}
    \begin{spacing}{1.1}
    \begin{algorithmic}[1]
    \Procedure{TrainStep}{$\bfx(0),{\{\bfmu^l\}}_{l=1}^L,c, D_{\theta}$}
    \State \textbf{sample} $\ln (\sigma) \sim \matN(P_{\text{mean}},P_{\text{std}}^2)$
    \State $\sigma \gets \exp\bleftBrackets\ln(\sigma)\brightBrackets$
    
    \For{$l \in \{1,2,\cdots, L\}$}
        \State \textbf{sample} $\bfn^l \sim \matN(0,\bfI)$
        \State $\bftilx_0^l(t)\gets\bfx(0) + \alpha\sigma\bfmu^l, \bftilx^l(t) \gets \bftilx_0^l(t) + \sigma\bfn^l$
    \EndFor
    \State $\hat \bfx(0) \gets D_{\theta}\leftBrackets \{\bftilx^l(t)\}_{l=1}^L;\sigma;c\rightBrackets$ \Comment{\cref{eq:relation_D_and_F}}
    \State Take gradient descent step on
    \State \quad$\nabla_{\bfx} \bbE_{\sigma,\bfx(0),\bfn}\leftSquareBrackets{\lambda(\sigma) \| \hat\bfx(0) -  \bfx(0)\|}_2^2\rightSquareBrackets$ \Comment{\cref{eq:all_training_loss}}
    \EndProcedure
    \end{algorithmic}
    \end{spacing}
\end{algorithm}
}
\newcommand{\sampleCode}{
\begin{algorithm}[t]
    \footnotesize
    \caption{Our stochastic sampler with $s(t) = {1}/{\lb1+\alpha t\rb}$ and $\sigma(t)=t$.}
    \label{alg:test}
    \begin{spacing}{1.1}
    \begin{algorithmic}[1]
    \Procedure{StochasticSampler}{${\{\bfmu^l\}}_{l=1}^L,c,D_{\theta}$}
        \For {$l\in \{1,2,\cdots,L\}$} 
            \State \textbf{sample} $\bfx_0^l \sim \matN(\alpha\sigma\bfmu^l,\sigma^2\bfI)$ 
        \EndFor
        \For {$i\in \{0,1,\cdots,N-1\} $} 
            \State $\gamma_i \gets \Schurn/N$ {\bf if} $t_i \in [\Stmin, \Stmax]$ {\bf else} 0
            \State $\hatt_i \gets t_i + \gamma_i t_i$
            \For {$l\in \{1,2,\cdots,L\}$} 
                \State \textbf{sample} $\bfeps_i^l \in \matN \leftBrackets0, S^{2}_{\text{noise}}\bfI\rightBrackets $
                \State $\hatbfx_i^l \gets \bfx_i^l + \alpha(\hatt_i - t_i) \bfmu^l +\sqrt{\hatt_i^2 - t_i^2}\bfeps_i^l$ \Comment{\cref{eq:enlarge_noise_and_mean}}
            \EndFor
            \For {$l\in \{1,2,\cdots,L\}$} 
                \State $\bfd_{i}^l \gets 
                \lb{\hatbfx_i^l - D_{\theta}\bleftBrackets \{\hatbfx_i^l\}_{l=1}^L;\sigma;c\brightBrackets}\rb/{\hatt_i}$ \Comment{\cref{eq:backward_denoiser_ode}}
                \State $\bfx_{i+1}^l\gets \hatbfx_{i}^l +(t_{i+1}-\hatt_{i}) \bfd_{i}^l$ 
            \EndFor
        \EndFor
        \State $\bfx_N \gets \text{mean}\leftBrackets\{\bfx_{N}^l\}_{l=1}^L\rightBrackets$
        \State  \Return $\bfx_{N}$
    \EndProcedure
    \end{algorithmic}
    \end{spacing}
\end{algorithm}
}

\newcommand{\comparisonTab}{
\begin{table}[t]
    {
        \caption{Quantitative results on (a) CUHK-CR1, (b) CUHK-CR2, (c) SEN12MS-CR, and (d) Sen2\_MTC\_New datasets. The metrics align with those used in prior studies on these datasets. The symbols $\uparrow$/$\downarrow$ indicate that higher/lower values correspond to better performance. The best results are highlighted in \textcolor[rgb]{1,0,0}{red} \textbf{bold} \underline{underline}, while the second-best results are marked in \textcolor[rgb]{0,0,1}{blue} \textbf{bold}. Dashed lines separate diffusion-based approaches from others.}
        \label{tab:all}
    }
    \vspace{-2mm}
    \resizebox{1\columnwidth}{!}{
        \setlength{\tabcolsep}{1mm}{
            \genCuhkTab
        }
    }
    \resizebox{1\columnwidth}{!}{
        \setlength{\tabcolsep}{1.2mm}{
            \begin{tabular}{@{}l|c c c c@{}}
                \Xhline{0.8pt}
                {\bf (c) SEN12MS-CR} & PSNR$\uparrow$ & SSIM$\uparrow$ & MAE$\downarrow$ & SAM$\downarrow$\\ 
                \hline
            McGAN~\cite{enomoto2017filmy} &25.14&0.744&0.048&15.676\\
            SAR-Opt-cGAN~\cite{grohnfeldt2018conditional}&25.59&0.764&0.043&15.494\\
            SAR2OPT~\cite{bermudez2018sar}& 25.87&0.793&0.042&14.788\\
            SpA GAN~\cite{pan2020cloud}&24.78&0.754&0.045&18.085\\
            Simulation-Fusion GAN~\cite{gao2020cloud}&24.73&0.701&0.045&16.633\\
            DSen2-CR~\cite{meraner2020cloud}&27.76&0.874&0.031&\s9.472\\
            GLF-CR~\cite{xu2022glf}&28.64&0.885&0.028&\s8.981\\
            UnCRtainTS L2~\cite{ebel2023uncrtaints}&28.90&0.880&0.027&\s8.320\\
            ACA-Net~\cite{huang2024attentive}& 29.78 & 0.896&0.025&\s7.770 \\
            \hdashline
            DiffCR~\cite{zou2024diffcr}&\bluebf{31.77}&\bluebf{0.902}&\bluebf{0.019}&\s\bluebf{5.821}\\
            Ours (EMRDM)&\redulbf{32.14}&\redulbf{0.924}&\redulbf{0.018}&\s\redulbf{5.267}\\
            \Xhline{0.4pt}
            \end{tabular}
        }
    }
    \vspace{0.2cm}
    \resizebox{1\columnwidth}{!}{
        \setlength{\tabcolsep}{2.4mm}{
            \begin{tabular}{@{}l|ccc}
                \Xhline{0.8pt}
                {\bf (d) Sen2\_MTC\_New} & PSNR$\uparrow$ & SSIM$\uparrow$ & LPIPS$\downarrow $ \\
                \hline
                McGAN~\cite{enomoto2017filmy}&17.448&0.513&0.447\\
                Pix2Pix~\cite{isola2017image-to-image}&16.985 &0.455&0.535\\
                AE~\cite{sintarasirikulchai2018a}&15.100&0.441&0.602\\
                STNet~\cite{chen2020thick}&16.206&0.427&0.503\\
                DSen2-CR~\cite{meraner2020cloud}&16.827&0.534&0.446\\
                STGAN~\cite{sarukkai2020cloud}&18.152&0.587&0.513\\
                CTGAN~\cite{huang2022ctgan}&18.308&0.609&0.384\\
                SEN12MS-CR-TS Net
                ~\cite{ebel2022sen12ms}&18.585&0.615&0.342\\
                PMAA~\cite{zou2023pmaa}&18.369&0.614&0.392\\
                UnCRtainTS~\cite{ebel2023uncrtaints}&18.770&0.631&0.333\\
                \hdashline
                DDPM-CR~\cite{jing2023denoising}&18.742&0.614&0.329\\
                DiffCR~\cite{zou2024diffcr}&\bluebf{19.150}&\bluebf{0.671}&\bluebf{0.291}\\
                Ours (EMRDM)&\redulbf{20.067}&\redulbf{0.709}&\redulbf{0.255}\\
                \Xhline{0.4pt}
            \end{tabular}}
    }
    \vspace{-2mm}
\end{table}
}

\newcommand{\genTrainingTable}{%
\toprule
{\bf Training configuration} &PSNR$\uparrow$&SSIM$\uparrow$&MAE$\downarrow$&SAM$\downarrow$&LPIPS$\downarrow$ \\
\midrule
\confhdr{a} \low{Baseline ($s(t)=1$)}&12.81& 0.342&0.204&13.005&0.718\\
\confhdr{b} \low{+ Corrupted images}&18.26&0.649&0.109&\s6.526&0.311\\
\confhdr{c} \low{+ IR images}&19.31&0.677&0.095&\s6.547&0.279\\
\confhdr{d} \low{+ Our MRDM framework}&\textbf{19.52}&0.679&\textbf{0.092}&\s6.551&0.278\\
\confhdr{e} \low{+ Our preconditioning}&19.47&\textbf{0.693}&0.093&\textbf{\s6.390}&\textbf{0.267}\\
\bottomrule
}

\newcommand{\abalationModuleTab}{%

\tabulinesep=0.7mm%
\tabulinestyle{0.17mm}%
\begin{table}[t]%
    \centering%
    \footnotesize%
    \caption{
    We conducted an ablation study on the Sen2\_MTC\_New dataset to evaluate our method by incrementally adding modules.
    }
    \label{tab:abalation}
    \vspace{-2mm}%
    \resizebox{\columnwidth}{!}{%
        \begin{tabu}{@{\ \ }l@{\ \ }|c@{\cs}c@{\cs}c@{\cs}c@{\cs}c}
            \genTrainingTable
        \end{tabu}%
    }%
    \vspace*{-2mm}%
\end{table}%
}

\newcommand{\abalationParamTab}{
\begin{table}
    \centering
    \caption{Hyperparameter analysis on the Sen2\_MTC\_New dataset.}
    \label{tab:hyperparameter}
    \small
    \vspace{-2mm}
    \resizebox{\columnwidth}{!}{%
    \begin{tabular}{ccc|ccccc
                    }
         \Xhline{0.8pt}
         \multicolumn{3}{c|}{Configurations} & \multicolumn{5}{c}{Metrics} \\
         \Xhline{0.4pt}
         $\alpha$ & $\sigmatmax$ & $N$ & PSNR$\uparrow$ & SSIM$\uparrow$ & MAE$\downarrow$ & SAM$\downarrow$ & LPIPS$\downarrow$ \\
        \Xhline{0.4pt}  
        0.2 & \multirow{6}*{100.0} & \multirow{6}*{5} & 19.34& 0.692&0.095 & 6.306&0.269\\
        0.5 & &&19.14&0.675&0.097&6.580&0.283\\
        0.8 & &&19.90&0.689&0.088&6.249&0.260\\
        1.0 & &&19.44&0.688&0.091&6.367&0.273\\
        2.0 & &&19.77&0.704&0.087&5.922&0.262\\
        3.0 & &&{\bf 20.00}&{\bf 0.708}&{\bf 0.084}&{\bf 5.710}&{\bf 0.255}\\
        4.0 & &&19.76&0.695&0.087&5.821&0.263\\
        \hline \hline
        \multirow{6}*{3.0} & 40 & \multirow{6}*{5}&19.58&0.701&0.087&5.764&0.260\\
         & 60 &&19.88&0.706&0.085&5.733&0.257\\
         & 80 &&19.96&0.707&{\bf 0.084}&5.726&0.256\\
         & 100&&20.00&{\bf 0.708}&{\bf 0.084}&{\bf 5.710}&{\bf 0.255}\\
         & 150&&{\bf 20.03}&0.707&0.085&5.730&0.256\\
         & 200&&{\bf 20.03}&0.707&0.085&5.723&0.256\\
         & 300&&20.02&0.705&0.086&5.728&0.257\\
        \hline \hline
        \multirow{6}*{3.0} & \multirow{6}*{100} & 4 & 19.98&0.702&0.085&5.744&0.259\\
        &  &5&{\bf 20.00}&{\bf 0.708}&{\bf 0.084}&5.710&{\bf 0.255}\\
        &  &6&19.97&0.705&0.084&5.710&0.257\\
        &  &8&19.89&0.700&0.085&{\bf 5.695}&0.257\\
        &  &10&19.89&0.700&0.085&{\bf 5.695}&0.257\\
        &  &15&19.55&0.672&0.088&5.715&0.261\\
        &  &50&19.19&0.641&0.091&5.857&0.270\\
         \Xhline{0.4pt}
    \end{tabular}
    }
    \vspace{-2mm}
\end{table}
}

\newcommand{\figRelation}{
\begin{figure}
    \centering
    \includegraphics[width=\columnwidth]{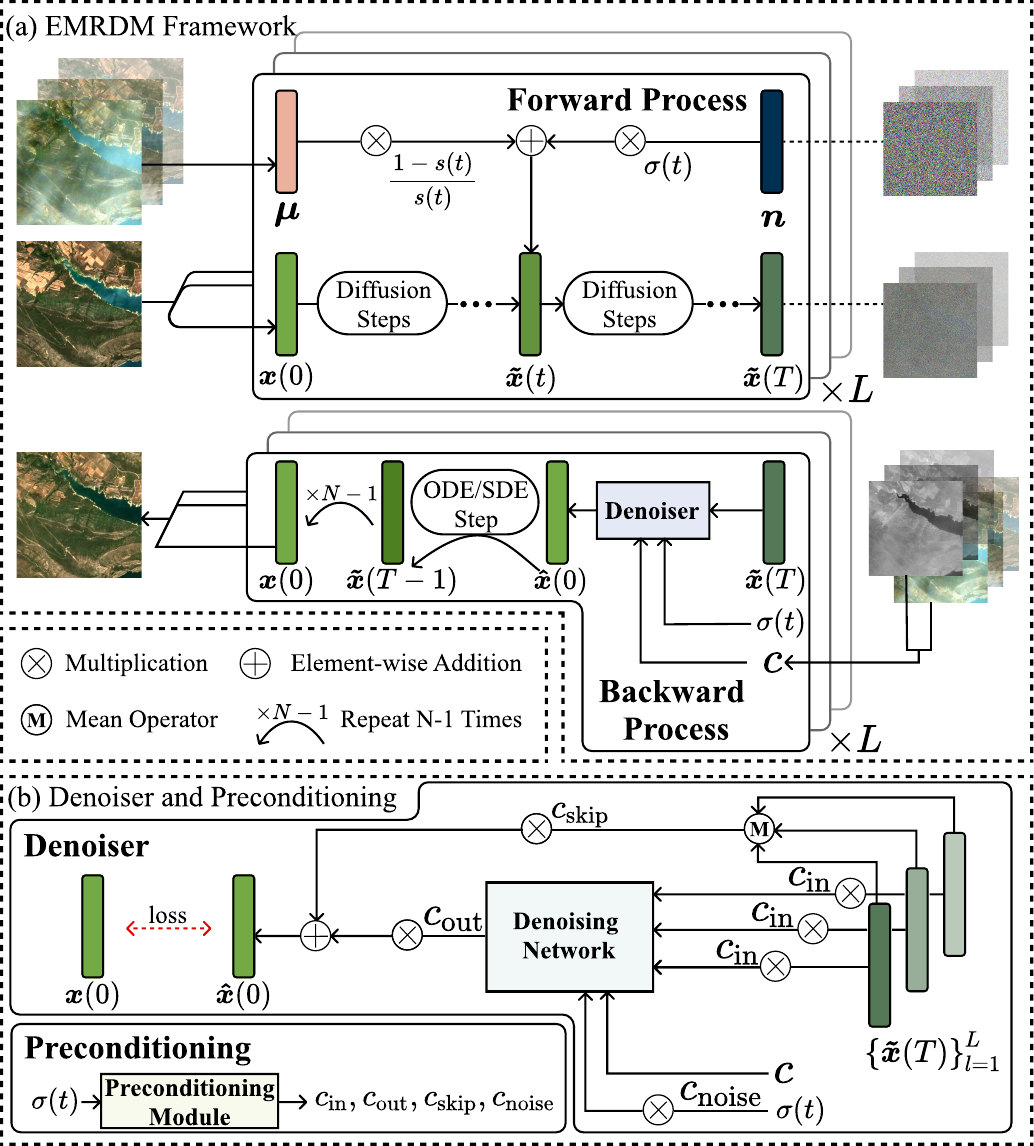}
    \vspace{-2mm}
    \caption{(a) The EMRDM framework comprises a forward process and a backward process that contains a denoiser. (b) The denoiser consists primarily of a denoising network, where the preconditioning module generates reparameterized factors $\cins,\couts,\cskips,\cnoises$ based on noise level $\sigma(t)$. We show the multi-temporal condition with the sequence length $L$.}
    \label{fig:relationship}
    \vspace{-2mm}
\end{figure}
}

\newcommand{\figVisual}{
    \begin{figure*}
        \centering
        \includegraphics[width=\linewidth]{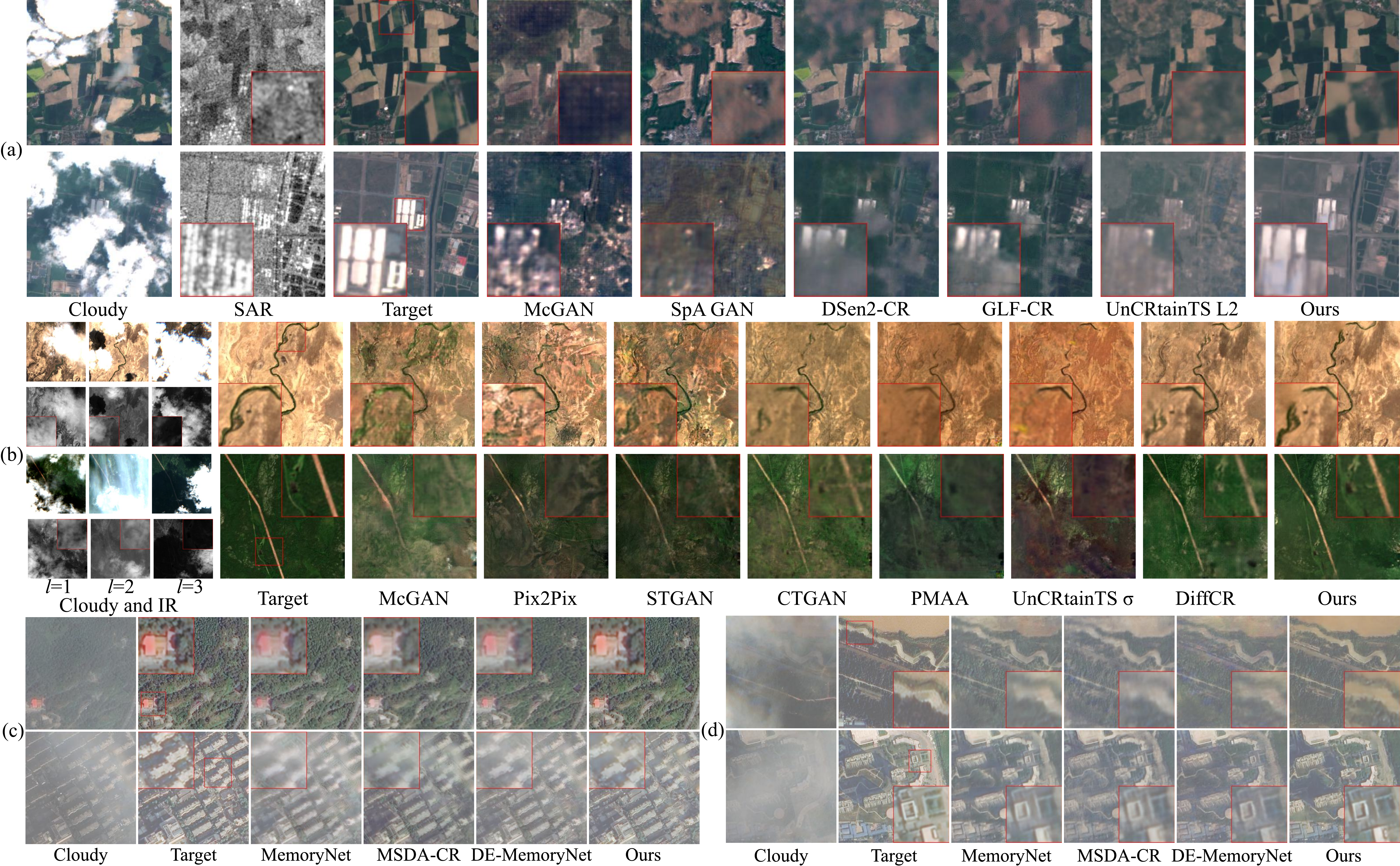}
        \vspace{-2mm}
        \caption{
        (a) SEN12MS-CR dataset results: RGB channels for optical imagery (linearly enhanced for visualization) and VV channel for SAR imagery. 
        GLF-CR results are obtained by combining four separately processed subimages as it processes $128\times128$ images ($256\times256$ for others).
        (b) Sen2\_MTC\_New dataset results. (c,d) RGB channel results on CUHK-CR1 and CUHK-CR2 datasets, respectively.
        }
        \label{fig:visual}
        \vspace{-2mm}
    \end{figure*}
}

\newcommand{\abalationFig}{
\begin{figure}
  \centering
  \includegraphics[width=\columnwidth]{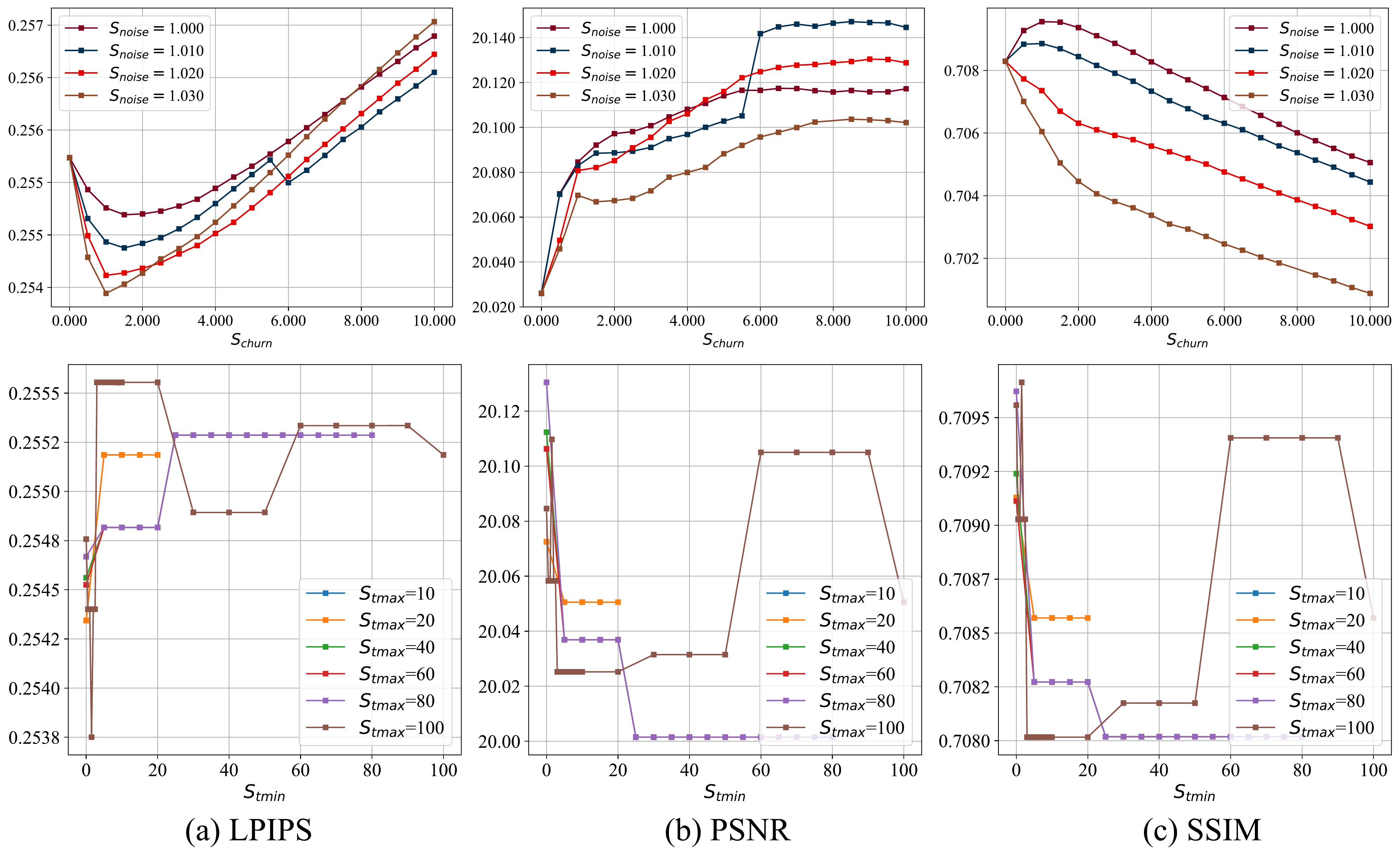}
  \caption{Analysis of our samplers on the Sen2\_MTC\_New dataset. When $\Schurn=0$, the sampler reduces to be deterministic. 
  The upper row shows the effects of $\Schurn$ and $\Snoise$ by fixing $\Stmin=0$ and $\Stmax\ge100$. The lower row examines the effects of $\Stmin$ and $\Stmax$ with fixed $\Schurn=1$ and $\Snoise=1$. Note that $\Stmin>=\Stmax$ is excluded as this leads to a deterministic sampler.}
  \label{fig:abalation_stochastic}
  \vspace{-2mm}
\end{figure}
}
\newcommand{\abalationAttnFig}{
\begin{figure}[t]
  \centering
  \includegraphics[width=\columnwidth]{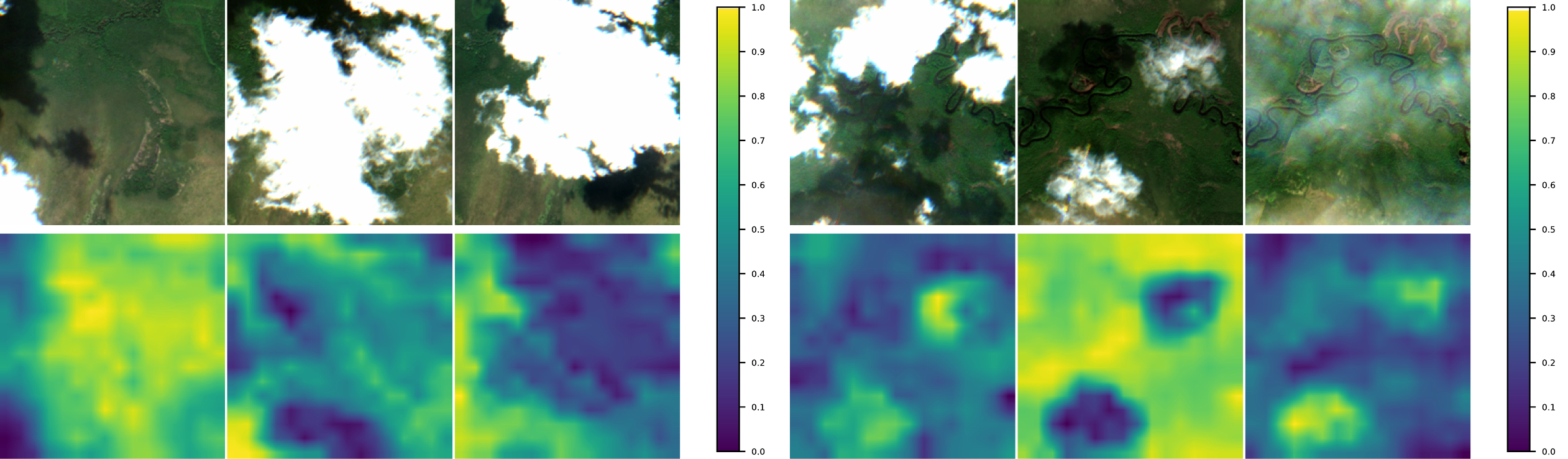}
  \vspace{-2mm}
  \caption{
  Visualizations of attention masks and their corresponding cloudy images from two cases on the Sen2\_MTC\_New dataset. Each mask at different time points is normalized to the range $[0, 1]$ and upsampled using bilinear interpolation to match the size of the cloudy images for clarity. 
  The left panel shows a case from head $0$, while the right panel displays a case from head $15$.
  }
  \label{fig:abalation_attn}
  \vspace{-2mm}
\end{figure}
}

\newcommand{\figFramework}{
\begin{figure}
    \centering
        \includegraphics[width=1\columnwidth]{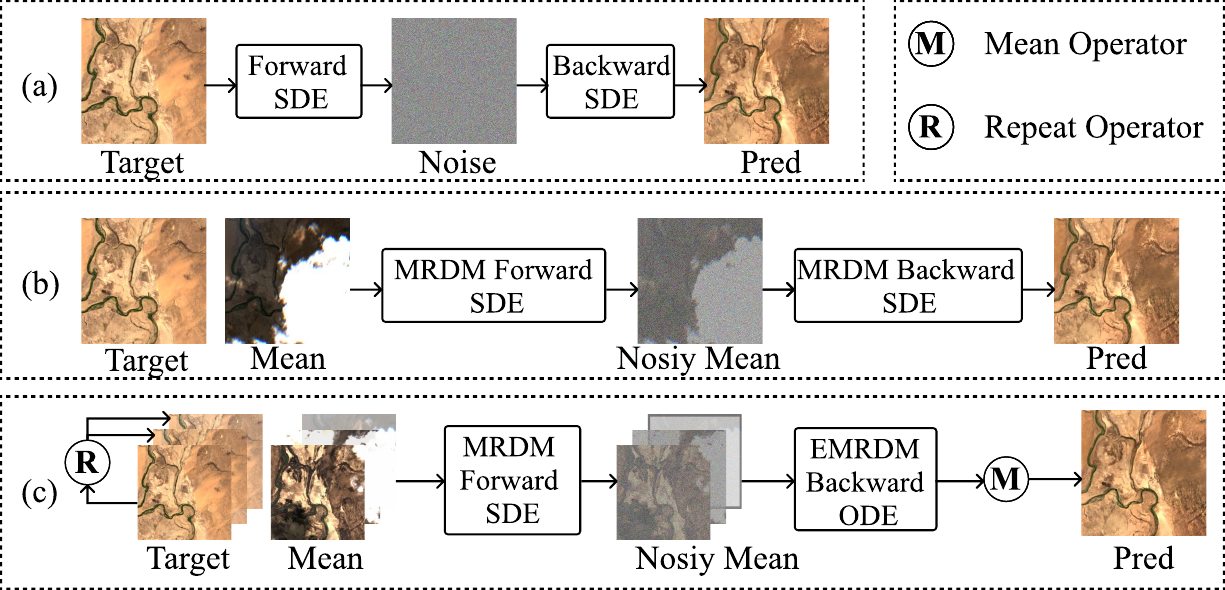}
        \vspace{-2mm}
        \captionsetup{type=figure}
        \captionof{figure}{
            Comparison of EMRDM (c) with generative DMs (a) and  MRDMs (b). Here, \textit{target} is the cloudless image, \textit{pred} is the CR prediction result, \textit{mean} is the cloudy image,  and \textit{noisy mean} is the noisy cloudy image. The forward processes of (a), (b), and (c) generate diffused images approximated by \textit{noise} (for DMs) and \textit{noisy mean} (for EMRDM and MRDMs), respectively.
        }
        \label{fig:framework}
        \vspace{-2mm}
\end{figure}
}
\newcommand{\scorefunc}{s_{\theta}\blb \bftxt\brb}

\newtheorem{assumption}{Assumption}[section]

\newcommand{\genCuhkTab}{
    \begin{tabular}{@{}l|ccc|ccc@{}}
        \Xhline{0.8pt}
        \multirow{2}*{Method}&
        \multicolumn{3}{c|}{\textbf{(a) CUHK-CR1}}&\multicolumn{3}{c}{\textbf{(b) CUHK-CR2}}\\
&PSNR$\uparrow$&SSIM$\uparrow$&LPIPS$\downarrow$&PSNR$\uparrow$&SSIM$\uparrow$&LPIPS$\downarrow$\\
        \hline
            SpA-GAN~\cite{pan2020cloud}   & 20.999 & 0.5162 & 0.0830 & 19.680 & 0.3952 & 0.1201\\
        AMGAN-CR~\cite{xu2022attention}  & 20.867 & 0.4986 & 0.1075 & 20.172 & 0.4900 & 0.093\\
            CVAE~\cite{ding2022uncertainty}      & 24.252 & 0.7252 & 0.1075 & 22.631 & 0.6302 & 0.0489\\
        MemoryNet~\cite{zhang2023memory} & 26.073 & 0.7741 & 0.0315 & 24.224 & 0.6838 & 0.0403\\
        MSDA-CR~\cite{yu2022cloud}   & 25.435 & 0.7483 & 0.0374 & 23.755 & 0.6661 & 0.0433\\
            \hdashline
            DE-MemoryNet~\cite{sui2024diffusion} & \bluebf{26.183} & \bluebf{0.7746} & \bluebf{0.0290} & \bluebf{24.348} & \bluebf{0.6843} &\bluebf{ 0.0369}\\
            DE-MSDA~\cite{sui2024diffusion}   & 25.739 & 0.7592 & 0.0321 & 23.968 & 0.6737 & 0.0372\\
            Ours (EMRDM) & \redulbf{27.281} &\redulbf{0.8007}& \redulbf{0.0218}&\redulbf{24.594}&\redulbf{0.6951}&\redulbf{0.0301}\\
        \Xhline{0.4pt}
    \end{tabular}
}

\newcommand{\genSeqLengthTable}{%
\toprule
{\bf Sequence Length} &PSNR$\uparrow$&SSIM$\uparrow$&MAE$\downarrow$&SAM$\downarrow$&LPIPS$\downarrow$ \\
\midrule
    $L=1$&16.09&0.493&0.146&7.773&0.440\\
    $L=2$&18.10&0.623&0.106&7.313&0.344\\
    $L=3$&{\bf 20.07}&{\bf 0.709}&{\bf 0.084}&{\bf 5.670}&{\bf 0.255}\\
\bottomrule
}

\newcommand{\abalationSeqLengthTab}{%
\tabulinesep=0.7mm%
\tabulinestyle{0.17mm}%
\begin{table}[t]%
    \centering%
    \footnotesize%
    \caption{
    Analysis of $L$ 
    on the Sen2\_MTC\_New dataset. 
    }
    \label{tab:sequence_length}
    \vspace{-2mm}%
    \resizebox{\columnwidth}{!}{%
        \begin{tabu}{@{\ \ }l@{\ \ }|c@{\cs}c@{\cs}c@{\cs}c@{\cs}c}
            \genSeqLengthTable
        \end{tabu}%
    }%
    \vspace*{-2mm}%
\end{table}%
}

\newcommand{\dsampleCode}{
\begin{algorithm}[t]
    \caption{Our stochastic sampler with arbitrary $\st$ and $\sigmat$.}
    \label{alg:dtest}
    \begin{spacing}{1.1}
    \begin{algorithmic}[1]
    \Procedure{StochasticSampler}{$D_{\theta},\{\bfmu^l\}_{l=1}^L,c$}
        \For {$l\in \{1,2,\cdots,L\}$} \Comment{Individually sample the initial state for $L$ time points} 
            \State \textbf{sample} $\bfx_0^l \sim \matN(\cfrac{1-s(t_0)}{s(t_0)}\bfmu^l,\sigma(t_0)^2\bfI)$ \Comment{$\bfx_0^l$ is a noisy corrupted image}
        \EndFor
        \For {$i\in \{0,1,\cdots,N-1\} $} \Comment{Repeat the sampling step $N$ times}
            \If{$t_i \in [\Stmin, \Stmax] $}
                \Comment{$[\Stmin,\Stmax]$ define the stochastic sampling range}
                \State $\gamma_i \gets \cfrac{\Schurn}{N}$ \Comment{$\Schurn$ and $N$ determine $\gamma_i$} 
            \Else \Comment{For $t_i$ outside the range $[\Stmin, \Stmax]$, use deterministic sampling}
                \State $\gamma_i \gets 0$ \Comment{Setting $\gamma_i=0$ leads to deterministic sampling}
            \EndIf
            \State $\hatt_i \gets t_i + \gamma_i t_i$
            \Comment{$\gamma_i$ regulates the extent of stochastic perturbation}
            \For {$l\in \{1,2,\cdots,L\}$} 
            \Comment{Individually perform denoising for $L$ time points}
                \State \textbf{sample} $\bfeps_i \in \matN \lb0, \Snoise^2\bfI\rb $ 
                \Comment{Sample the noise for stochastic perturbation}
                \State $\hatxil \gets \bfx_i^l + \lb
                    \cfrac{1-s(\hatt)}{s(\hatt_i)} - \cfrac{1-s(t_i)}{s(t_i)}
                \rb \bfmu^l +\sqrt{{\sigma(\hatt_i)}^2 - {\sigma(t_i)}^2}\bfeps_i$ \Comment{Use \cref{eq:enlarge_noise_and_mean} for stochastic perturbation}
            \EndFor
            \For {$l\in \{1,2,\cdots,L\}$} \Comment{Individually take Euler step for $L$ time points}
                \State $\bfd_{i}^l \gets - \cfrac{\dot s(\hatti)}{{s(\hatti)}^2}\bfmu - \cfrac{\dot\sigma(\hatti)}{\sigma(\hatti)} \lsb
                    \Dt\lb 
                        {\lB \hatxil \rB}_{l=1}^L;\sigma(\hatti);c
                    \rb+\cfrac{1-s(\hatti)}{s(\hatti)}\bfmu - \hatxil
                \rsb$ 
                \Comment{Use \cref{eq:backward_denoiser_ode}}
            \State $\bfx_{i+1}^l\gets \hatbfx_{i}^l +(t_{i+1}-\hatt_{i}) \bfd_{i}^l$ 
            \Comment{Take an Euler step from $\hatti$ to $t_{i+1}$}
            \EndFor
        \EndFor
        \State $\bfx_N = \text{mean}\leftBrackets\{\bfx_{N}^l\}_{l=1}^L\rightBrackets$
        \Comment{Use the mean operator to collapse the temporal dimension and calculate the final result}
        \State  \Return $\bfx_{N}$
        \Comment{The result is a single restored image}
    \EndProcedure
    \end{algorithmic}
    \end{spacing}
\end{algorithm}
}

\newcommand{\samplersFig}{
    \begin{figure*}[t]
    \centering
    \includegraphics[width=\linewidth]{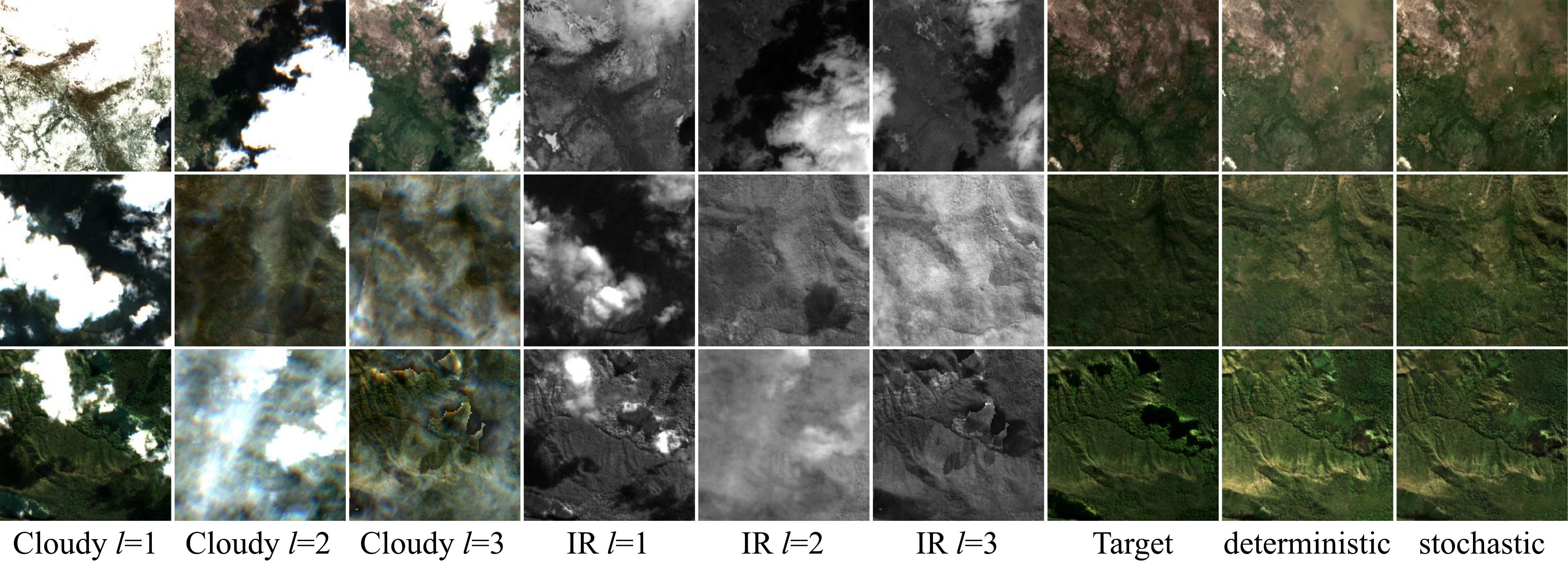}
    \caption{Visual results generated by the stochastic sampler and the deterministic sampler. For the deterministic sampler, we set $N=5$, $\sigmatmin=0.001$ and $\sigmatmax=100$. For the stochastic sampler, we set $N=5$, $\sigmatmin=0.001$, $\sigmatmax=100$, $\Schurn=1.0$, $\Snoise=1.0$, $\Stmin=0$ and $\Stmax=100$.
    }
    \label{app:fig:samplers}
    \end{figure*}
}

\newcommand{\cuhkAppFig}{
    \begin{figure*}[t]
    \centering
    \includegraphics[width=\linewidth]{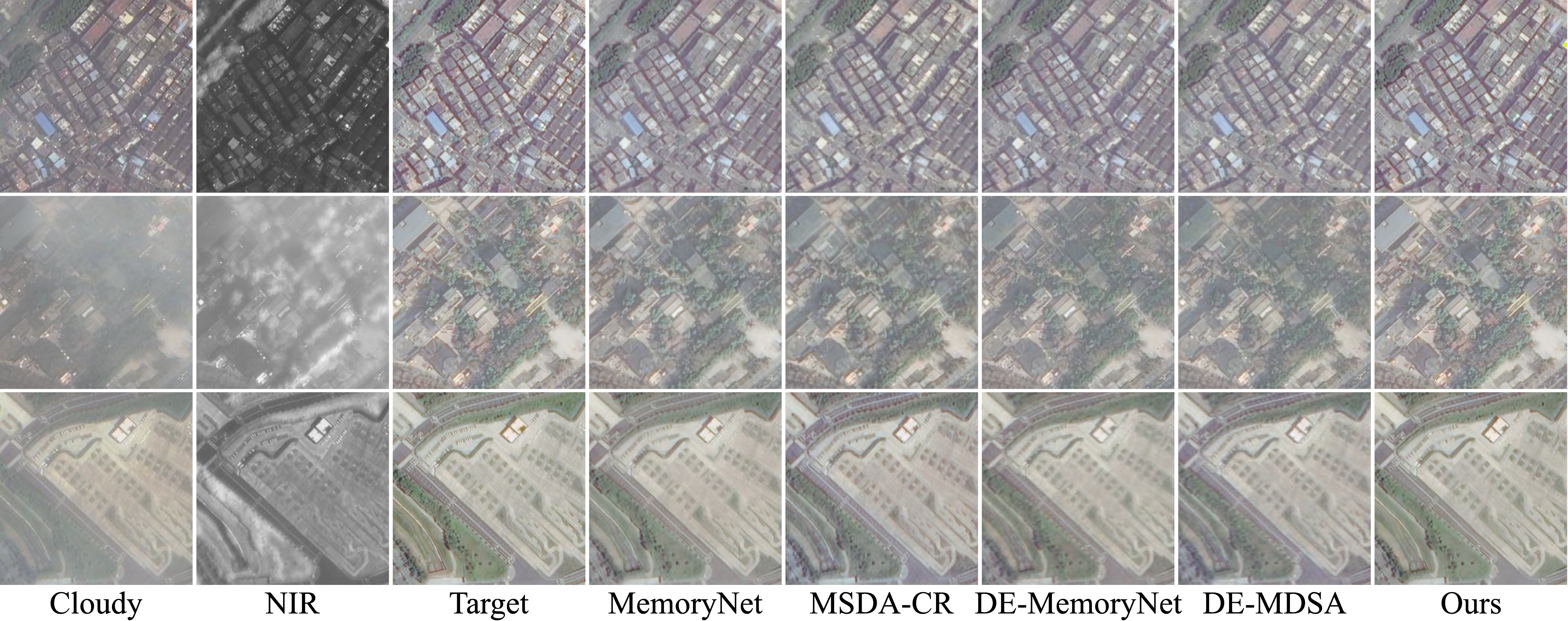}
    \caption{Additional visual results on the CUHK-CR1 dataset.}
    \label{app:fig:cuhk-cr1}
    \end{figure*}
}

\newcommand{\cuhkVTAppFig}{
    \begin{figure*}[t]
    \centering
    \includegraphics[width=\linewidth]{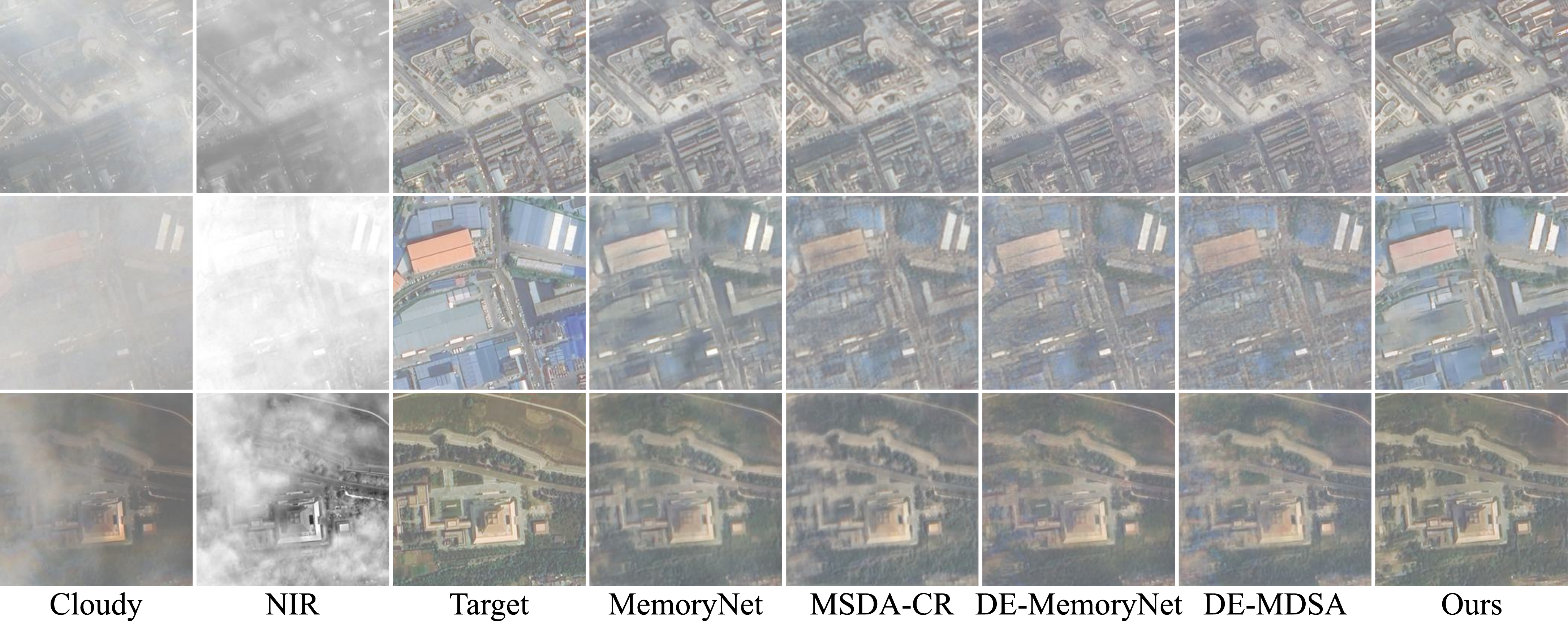}
    \caption{Additional visual results on the CUHK-CR2 dataset.}
    \label{app:fig:cuhk-cr2}
    \end{figure*}
}

\newcommand{\SENMSCRAppFig}{
    \begin{figure*}[t]
    \centering
    \includegraphics[width=\linewidth]{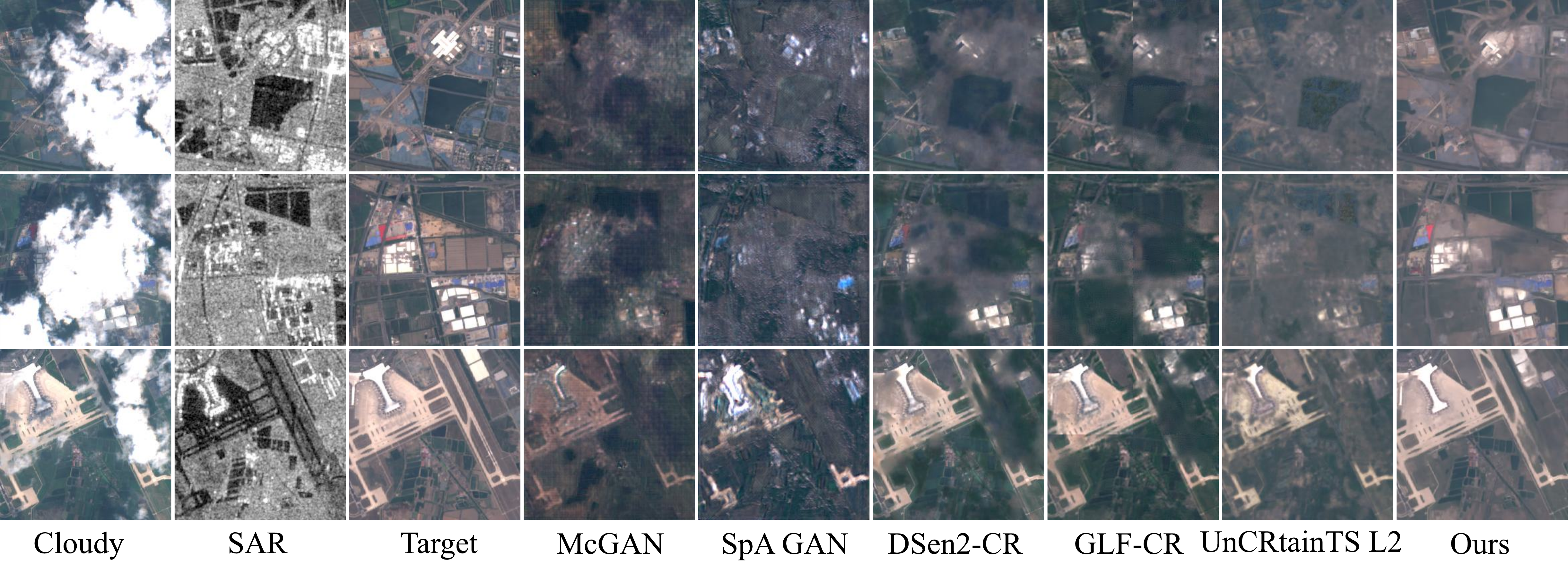}
    \caption{Additional visual results on the SEN12MS-CR dataset. As GLF-CR~\cite{xu2022glf} can only process $128\times128$ images, unlike others ($256\times256$), we divide each image into four parts, process them individually, and merge the results. Optical image brightness is linearly enhanced for visualization.}
    \label{app:fig:sen12mscr}
    \end{figure*}
}

\newcommand{\SenMTCNewAppFig}{
    \begin{figure*}[t]
    \centering
    \includegraphics[width=\linewidth]{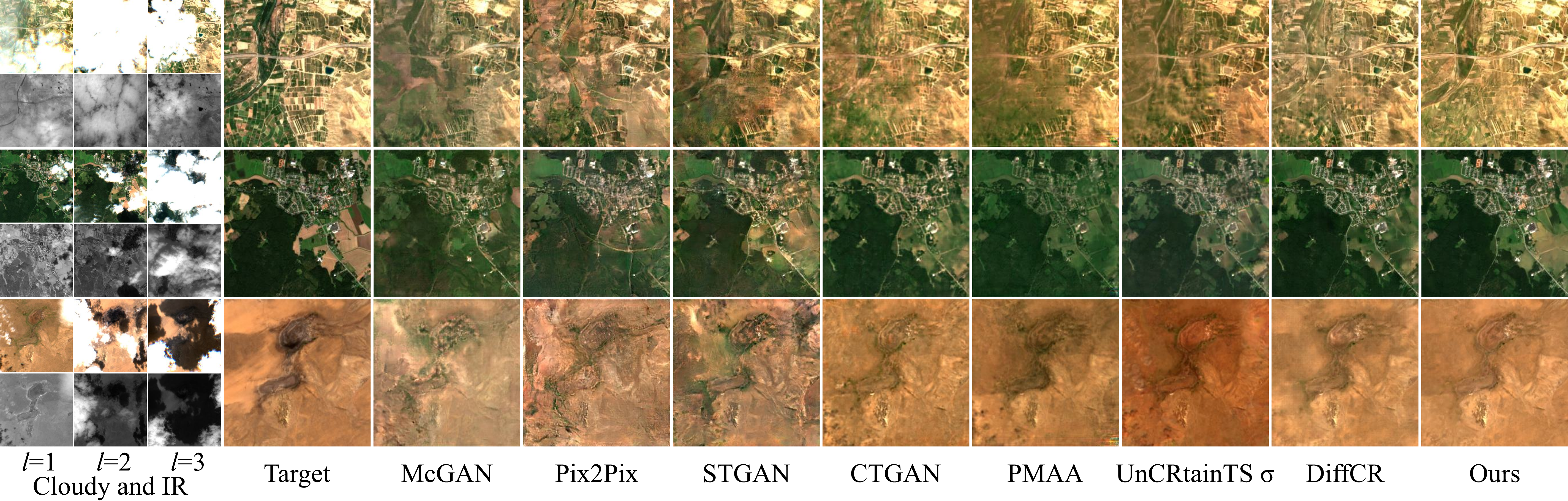}
    \caption{Additional visual results on the Sen2\_MTC\_New dataset.}
    \label{app:fig:senmtcnew}
    \end{figure*}
}

\newcommand{\hyperparamsFig}{
    \begin{figure*}[t]
    \centering
    \includegraphics[width=\linewidth]{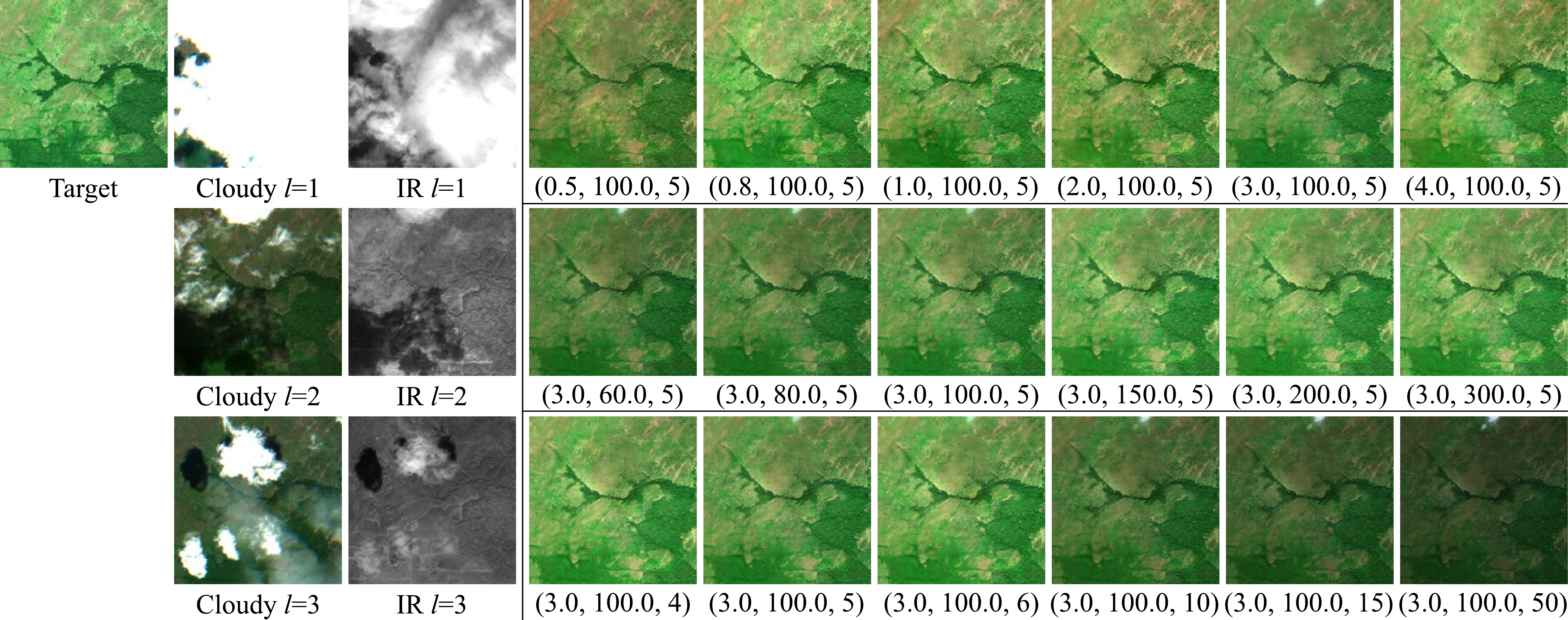}
    \caption{Visual results under different configurations of ($\alpha$, $\sigma_{\text{max}}$, $N$). For example, $(3.0, 100.0, 5)$ represents the restored results with $\alpha=3.0$, $\sigmatmax=100.0$ and $N=5$.}
    \label{app:fig:hyperparameters}
    \end{figure*}
}

\newcommand{\trainAllDetails}{
\begin{table*}[t]
    \centering
    \caption{Details of our best training and testing configurations.}
    \begin{adjustbox}{width=\linewidth}
    \begin{tabular}{lcccc}
        \toprule
        \textbf{} & \textbf{CUHK-CR1}& \textbf{CUHK-CR2}& \textbf{SEN12MS-CR}& \textbf{Sen2\_MTC\_New} \\
        \midrule
        Parameters & 39.13M & 39.13M & 39.13M& 148.88M\\
        \arrayrulecolor{black!50} \midrule \arrayrulecolor{black}
        Training Steps &22,500  &26,300 & 446,700   &64,141\\
        Training Epochs&500     &470    & 46      &500   \\
        Batch Size     &4 & 2 & 2&8\\
        Precision         & tf32    & tf32 & tf32  & tf32 \\
        Training Hardware & 3 RTX 3090 & 4 RTX 4090 & 4 RTX 4090   & 4 RTX 4090\\
        \midrule
        In Channels & 8 (= 4 + 0 + 4) & 8 (= 4 + 0 + 4) & 28 (= 13 + 2 + 13) &7 (= 3 + 1 + 3)\\
        Out Channels& 4 & 4 & 13 &3\\
        \arrayrulecolor{black!50} \midrule \arrayrulecolor{black}
        Patch Size  & 1 & 1 & 1 & 4\\
        Levels (Local + Global Attention) & 2 + 2 & 2 + 2 & 2 + 2& 2 + 1\\
        Depth & [2, 2, 2, 2] & [2, 2, 2, 2] & [2, 2, 2, 2] & [2, 2, 16]\\
        Widths & [128, 256, 384, 768] & [128, 256, 384, 768] & [128, 256, 384, 768]  & [256, 512, 768]\\
        FFN Intermediate Widths & [256, 512, 768, 1536]  & [256, 512, 768, 1536]   & [256, 512, 768, 1536]  & [512, 1024, 1536] \\
        Attention Heads (Width / Head Dim) & [2, 4, 6, 12] & [2, 4, 6, 12] & [2, 4, 6, 12] & [4, 8, 12] \\
        Attention Head Dim & 64 & 64 & 64 & 64\\
        Neighborhood Kernel Size & 7 & 7 & 7& 7 \\
        Dropout Rate & [0.0, 0.0, 0.0, 0.1] & [0.0, 0.0, 0.0, 0.1] & [0.0, 0.0, 0.0, 0.1] & [0.0, 0.0, 0.0, 0.0]\\
        \arrayrulecolor{black!50} \midrule \arrayrulecolor{black}
        Mapping Depth & 2 & 2 & 2 & 2\\
        Mapping Width & 768 & 768 & 768& 768 \\
        Mapping FFN Intermediate Width & 1536 & 1536 & 1536& 1536\\
        Mapping Dropout Rate& 0.1 & 0.1 & 0.1& 0.1\\
        \midrule
        $\alpha$    &3.0   &3.0   &3.0   &3.0\\
        $\sigmatd$  &1.0   &1.0   &1.0   &1.0\\
        $\sigmatmu$ &1.0   &1.0   &1.0   &1.0\\
        $\sigmacov$ &0.9   &0.9   &0.9   &0.9\\
        $P_{\text{mean}}$ in~\cref{alg:train}   &-1.4   &-1.2   &-1.2   &-1.4\\
        $P_{\text{std}} $ in~\cref{alg:train}   &1.4    &1.2    &1.2   &1.4\\
        \midrule
        Optimizer & AdamW & AdamW & AdamW & AdamW\\
        Learning Rate & 1e-4 & 1e-4 & 1e-4 & 1e-4\\
        Betas & [0.9, 0.999] & [0.9, 0.999] & [0.9, 0.999] & [0.9, 0.999]\\
        Eps & 1e-8 & 1e-8 & 1e-8 & 1e-8\\
        Weight Decay & 1e-2 & 1e-2 & 1e-2 & 1e-2\\
        \arrayrulecolor{black!50} \midrule \arrayrulecolor{black}
        EMA Decay & 0.9999 & 0.9999 & 0.9999& 0.9999\\
        \midrule
        Sampling Steps $N$ & 4 & 4 & 5 & 5\\
        $\sigmatmin$&0.001 &0.001 &0.001 &0.001\\
        $\sigmatmax$&100   &100   &100   &100\\
        $\Schurn$   & 0.1     & 2.5& 5.0 &1.0\\
        $\Snoise$   & 0.995   & 1.0& 1.023&1.0\\
        $\Stmin$    & 0.0     & 0.0& 0.0&0.0\\
        $\Stmax$    &100000000& 100000000& 100000000&100.0\\
        \bottomrule
    \end{tabular}
    \end{adjustbox}
    \label{tab:training_details}
\end{table*}
}

\newcommand{\parammacTab}{
\begin{table}[!t]
    \centering
    \caption{
    The Comparison of Params (the number of parameters) and MACs (multiply-accumulate operations). 
    }
    \label{tab:params_and_macs}
    \begin{minipage}[c]{0.49\columnwidth}
        \centering
        \resizebox{\textwidth}{!}{
            \begin{tabular}{@{}l|cccc@{}}
            \Xhline{0.8pt}
                {\bf (a) SEN12MS-CR}&GLF-CR  &  UnCRtainTS L2 & DiffCR & EMRDM  \\
                \hline
                Params (M) & 14.827 & 0.519 & 22.96    & 39.13 \\
                MACs (G)   & 245.28 & 28.02 & 29.37    & 83.57   \\
            \Xhline{0.4pt}
            \end{tabular}
        }
    \end{minipage}
    \begin{minipage}[c]{0.49\columnwidth}
        \centering
        \resizebox{\textwidth}{!}{
            \begin{tabular}{@{}l|cccc@{}}
            \Xhline{0.8pt}
                {\bf (b) CUHK-CR}   & MemoryNet  & MSDA-CR &  DE & EMRDM  \\
                \hline
                Params (M)      & 3.64       & 3.91    &  36.80   & 39.13  \\
                MACs (G)        & 548.65    & 53.45   &  199.15  & 83.33  \\
            \Xhline{0.4pt}
            \end{tabular}
        }
    \end{minipage}
    \vspace{2mm}
    \begin{minipage}[c]{0.98\columnwidth}
        \centering
        \resizebox{\textwidth}{!}{
            \begin{tabular}{@{}l|cccccccc@{}}
            \Xhline{0.8pt}
                {\bf (c) Sen2\_MTC\_New}  & STGAN & CTGAN & CR-TS Net& PMAA &  UnCRtainTS & DDPM-CR & DiffCR & EMRDM \\
                \hline
                Params (M)&231.93&642.92&38.68&3.45 &0.56& 445.44& 22.91& 148.88\\
                MACs (G) &1094.94
                &632.05
                &7602.97
                &92.35
                &37.16
                &852.37
                &45.86
                &74.39\\
            \Xhline{0.4pt}
            \end{tabular}
        }
    \end{minipage}
\end{table}
}
\usepackage[accsupp]{axessibility}  
\definecolor{cvprblue}{rgb}{0.21,0.49,0.74}
\usepackage[pagebackref,breaklinks,colorlinks,allcolors=cvprblue]{hyperref}

\def\paperID{***} 
\def\confName{CVPR}
\def\confYear{2025}

\title{Effective Cloud Removal for Remote Sensing Images by an Improved Mean-Reverting Denoising Model with Elucidated Design Space}

\author{%
  Yi Liu$^1$, Wengen Li$^{1\ast}$, Jihong Guan$^{1\ast}$, Shuigeng Zhou$^2$, Yichao Zhang$^1$ \\
  $^1$ Tongji University, $^2$ Fudan University\\
  {\tt \small \{liuyi61,lwengen,jhguan,yichaozhang\}@tongji.edu.cn, sgzhou@fudan.edu.cn} \\
}

\begin{document}

\maketitle
\makeatletter\def\Hy@Warning#1{}\makeatother \let\thefootnote\relax\footnotetext{\noindent${}^\ast$Corresponding author.}
\begin{abstract}
Cloud removal (CR) remains a challenging task in remote sensing image processing. 
Although diffusion models (DM) exhibit strong generative capabilities, their direct applications to CR are suboptimal, as they generate cloudless images from random noise, ignoring inherent information in cloudy inputs. 
To overcome this drawback, we develop a new CR model \textbf{EMRDM} based on mean-reverting diffusion models (MRDMs) to establish a direct diffusion process between cloudy and cloudless images. 
Compared to current MRDMs, EMRDM offers a modular framework with updatable modules and an elucidated design space, based on a reformulated forward process and a new ordinary differential equation (ODE)-based backward process. 
Leveraging our framework, we redesign key MRDM modules to boost CR performance, including restructuring the denoiser via a preconditioning technique, reorganizing the training process, and improving the sampling process by introducing deterministic and stochastic samplers. 
To achieve multi-temporal CR, we further develop a denoising network for simultaneously denoising sequential images. 
Experiments on mono-temporal and multi-temporal datasets demonstrate the superior performance of EMRDM. 
Our code is available at \url{https://github.com/Ly403/EMRDM}.
\end{abstract}    
\section{Introduction}
\label{sec:introduction}
\figFramework

Satellite imagery, as a fundamental remote sensing product~\cite{yuan2020deep,zhu2017deep}, enables diverse applications including environmental monitoring~\cite{vakalopoulou2015building}, land cover classification~\cite{kussul2017deep}, and agricultural monitoring~\cite{russwurm2017temporal}. 
However, cloud coverage severely affects the usability of satellite imagery.
Data analysis for the Moderate Resolution Imaging Spectroradiometer (MODIS) on the Terra and Aqua satellites indicates that about 67\% of the Earth's surface experiences cloud coverage~\cite{king2013spatial}.
Hence, cloud removal (CR) is a critical preliminary step in processing satellite imagery.

Recent advances in deep learning have driven the progress of CR~\cite{xiong2023sar-to-optical}, with generative adversarial networks (GANs)~\cite{goodfellow2014generative} becoming a predominant approach. 
However, the effectiveness of GANs in CR is undermined by training instability~\cite{salimans2016improved} and mode collapse~\cite{bau2019seeing}.
In comparison, diffusion models (DMs)~\cite{song2019generative,ho2020denoising,song2021scorebased} can overcome these limitations via enhanced training stability and output diversity, setting new benchmarks in image synthesis~\cite{dhariwal2021diffusion} and restoration~\cite{li2023diffusion}. 
Such advantages of DMs also extend to CR tasks~\cite{zou2024diffcr,zhao2023cloud,jing2023denoising,sui2024diffusion}.

Existing diffusion-based CR methods typically employ vanilla DM frameworks that start the diffusion process from pure noise (\cref{fig:framework} (a)).
However, this is unnecessary as cloudy images contain substantial unexploited information. 
Even worse, noise-initiated generation lacks pixel-level consistency, inducing distortion~\cite{blau2018perception} in restored images due to poor fine-grained controllability.
To resolve this, we propose the integration of mean-reverting diffusion models (MRDMs)~\cite{luo2023image} into CR. 
MRDMs start the diffusion process directly from noisy cloudy images (\cref{fig:framework} (b)), intrinsically preserving structural fidelity through pixel-level consistency constraints. 
Specifically, the forward process progressively diffuses the target image by injecting noise while maintaining the cloudy image as the distribution mean, yielding a noisy cloudy image (\textit{noisy mean}). 
Subsequent denoising in the backward process reconstructs the cloudless image (\textit{pred}) while preserving structural consistency. 
However, current MRDMs exhibit limitations due to their intricately coupled modules and opaque relationships among modules, impeding their application. 
Inspired by the successful designs of EDM~\cite{karras2022elucidating} in image generation, we conduct an in-depth analysis of the underlying mathematical principles of MRDMs to clarify the roles and interrelationships of modules within the MRDM framework. 
Based on these insights, we \textbf{E}lucidate the design space of \textbf{MRDM}s and propose a novel MRDM-based CR model, termed \textbf{EMRDM}. 
EMRDM offers a modular framework by reformulating the forward process through a stochastic differential equation (SDE) with simplified parameters and introducing an ordinary differential equation (ODE)-based backward process, as illustrated in~\cref{fig:framework} (c).
The framework offers two critical advantages: (1) an elucidated and flexible design space enabling orthogonal module modifications, and (2) seamless compatibility with generative DMs. 
Leveraging the advantages of our framework, we further redesign key MRDM modules to boost CR performance, focusing on the following enhancements:  
(1) We restructure the denoiser via a preconditioning technique, inspired by image generation methods~\cite{karras2022elucidating,consistencymodel}, to adaptively scale inputs and outputs of the denoising network according to noise levels. 
(2) We reorganize the training process and improve the sampling process. 
For practical sampling of CR results, we introduce novel deterministic and stochastic samplers based on the improved sampling process. 

To achieve multi-temporal CR, we further develop a denoising network that processes arbitrary-length image sequences. 
Specifically, for $L$ sequential cloudy images, our architecture employs $L$ weight-sharing encoders and bottleneck modules, compresses temporal features through a novel attention block, and reconstructs outputs via a single decoder. The generated attention masks are preserved and upsampled to various resolutions, serving as adaptive weights to fuse temporal skip feature maps. 
The preconditioning and training methods are modified to accommodate multi-temporal scenarios through sequential input compatibility optimization. 
During sampling, to ensure temporal restoration consistency, we independently restore each temporal instance under mono-temporal conditions and aggregate results through a mean fusion operator (\cref{fig:framework} (c)).


Our \textbf{contributions} are summarized as follows:

\begin{itemize}
\item [1)] 
We propose a novel CR model \textbf{EMRDM} 
that offers a modular framework with updatable modules and an elucidated design space. 
\item [2)] 
We develop a multi-temporal network with a temporal fusion method to denoise arbitrary-length image sequences. 
\item [3)]
We restructure the denoiser via a preconditioning method, improve training and sampling processes, and propose novel stochastic and deterministic samplers.
\item [4)] Experiments on mono-temporal and multi-temporal cases demonstrate the superior CR performance of EMRDM.
\end{itemize}
\section{Related Work}
\label{sec:relatedwork}

\noindent{\bf Cloud Removal.} 
CR methods are primarily divided into traditional methods~\cite{xu2019thin,xu2015thin,lin2012cloud} and deep learning-based methods, with the former offering better interpretability but generally inferior performance compared to data-driven methods. 
Deep learning-based methods are further categorized into mono-temporal~\cite{grohnfeldt2018conditional,enomoto2017filmy,bermudez2018sar,pan2020cloud,gao2020cloud,meraner2020cloud,xu2022glf,li2023transformer,ebel2023uncrtaints,zou2024diffcr} and multi-temporal~\cite{sarukkai2020cloud,huang2022ctgan,ebel2022sen12ms,ebel2023uncrtaints,zou2024diffcr,zhao2023cloud} paradigms based on single-image or sequential inputs.
%
Mono-temporal methods commonly employ vanilla conditional GANs (cGANs)~\cite{goodfellow2014generative,mirza2014conditional} in early applications~\cite{enomoto2017filmy,bermudez2018sar,grohnfeldt2018conditional}, with improvements including spatial attention~\cite{pan2020cloud} and transformer architectures~\cite{li2023transformer}.
Alternative frameworks include DMs~\cite{zou2024diffcr} and non-generative models~\cite{meraner2020cloud,xu2022glf,ebel2023uncrtaints}.
Multi-temporal strategies mainly use temporal cGAN~\cite{sarukkai2020cloud,huang2022ctgan}, temporal fusion attention (\eg, L-TAE~\cite{garnot2020lightweight,garnot2021panoptic}) as in~\cite{ebel2023uncrtaints}, and sequential DMs~\cite{zhao2023cloud}. 
CR methods are also classified as mono-modal~\cite{enomoto2017filmy,pan2020cloud,zou2024diffcr} or multi-modal, 
depending on the use of auxiliary modalities, including infrared (IR) and synthetic aperture radar (SAR) images.
Multi-modal methods involve modality concatenation~\cite{bermudez2018sar,grohnfeldt2018conditional,meraner2020cloud,ebel2023uncrtaints,ebel2022sen12ms,sarukkai2020cloud,huang2022ctgan} and specialized fusion modules~\cite{gao2020cloud,xu2022glf,zhao2023cloud}.

\noindent{\bf Diffusion Models.} 
Recent advances in generative modeling have witnessed DMs~\cite{ho2020denoising,song2019generative,song2021scorebased} surpass GANs~\cite{dhariwal2021diffusion} in image synthesis. 
Notable improvements to DMs~\cite{rombach2022high,karras2022elucidating,peebles2023scalable,crowson2024scalable} have also been proposed, with EDM~\cite{karras2022elucidating} and HDiT~\cite{crowson2024scalable} most crucial to our work.
EDM presents a framework that delineates the specific design decisions for DM components, while HDiT introduces an efficient hourglass diffusion transformer.
Inspired by the success of DMs in image generation, extensive studies have investigated their applications in image restoration~\cite{li2023diffusion}. These methods can be categorized as supervised learning~\cite{saharia2022image,saharia2022palette,xia2023diffir,luo2023image,luo2023refusion,delbracio2023inversion,liu2024residual,liu2023I2SB,yue2024resshift,bansal2024cold} or zero-shot learning~\cite{Choi_2021_ICCV,lugmayr2022repaint,kawar2022denoising,wang2023zeroshot,song2023pseudoinverse}. 
In the first category, several methods focus on generating images directly from noiseless or noisy corrupted images, such as IR-SDE~\cite{luo2023image}, InDI~\cite{delbracio2023inversion}, ResShift~\cite{yue2024resshift}, RDDM~\cite{liu2024residual}, and I2SB~\cite{liu2023I2SB}. Considering that starting with pure noise is inefficient, IR-SDE, InDI, ResShift, and RDDM all integrate the corrupted image and noise within the diffusion process. We extend this paradigm and apply it to CR.
\section{Methodology}
\label{sec:method}
As illustrated in~\cref{fig:relationship}, we introduce the EMRDM framework in~\cref{sec:framework}, propose a novel multi-temporal denoising network in~\cref{sec:network}, restructure the denoiser by the preconditioning technique in~\cref{sec:preconditioning}, and present our redesigned training and sampling process in~\cref{sec:sampling}. 
\subsection{Preliminary}
\label{sec:preliminary}
The forward process of DMs can be expressed as an SDE proposed by Song \etal (Eq. 5 in~\cite{song2021scorebased}), as follows:
\begin{equation}
    \ud\bfx  = \bff(\bfx,t)\ud t+g(t)\ud\bfomega_t,
  \label{eq:generating_forward_sde}
\end{equation}
where $\bfomega_t$ is a standard Brownian motion, $\bfx \in \bbR^d$ is an It\^o process, $\bff(\cdot,t): {\bbR}^d \rightarrow {\bbR}^d$ and $g(\cdot) : {\bbR}\rightarrow {\bbR}$ are the drift and diffusion coefficients, respectively, and $d$ is the dimensionality of images. Song \etal further derive a reverse probability flow ODE (Eq. 13 in \cite{song2021scorebased}) for sampling as
\begin{equation}
    \ud \bfx = \leftSquareBrackets\bff(\bfx,t) - \frac{1}{2}{g(t)}^2\nabla_{\bfx} \log p_t(\bfx)\rightSquareBrackets\ud t,
  \label{eq:generating_backward_ode}
\end{equation}
where $p_t(\bfx)$ is the probability density function (pdf) of $\bfx$ at time $t$. 
The score function $\score$ is predicted by a neural network. 
Therefore, the models proposed by \cite{song2021scorebased}, as well as our models, are \textit{score matching} models.

\figRelation
\subsection{The EMRDM Framework}
\label{sec:framework}
We reformulate the forward process of MRDMs to construct a stochastic process $\{\bfx(t)\}_{t=0}^T$ that transforms a target image into its noisy cloudy counterpart. The new ODE-based backward process iteratively denoises the corrupted images.

\noindent{\bf Forward Process.}
We transform the SDE in~\cref{eq:generating_forward_sde} into 
\begin{equation}
    \ud \bfx = f(t)(\bfx - \bfmu)\ud t + g(t)\ud\bfomega_t,
    \label{eq:restoration_forward_sde}
\end{equation}
where $\bfmu \in \bbR^d$ is the cloudy image, and the stochastic process $\bfx(t)$ is simplified to $\bfx$.  
According to~\cite{luo2023image}, 
\cref{eq:restoration_forward_sde} can be viewed as a special case of \cref{eq:generating_forward_sde} by defining $\bff(\bfx,t) = f(t)(\bfx - \bfmu)$. 
This setting yields a 
solution for the pdf of $\bfx(t)$ given $\bfx(0)$ and $\bfmu$: 
\begin{align}
    p_{0t}\bleftBrackets\bfx(t)\mid \bfx(0),\bfmu\brightBrackets&={s(t)}^{-d}\tilp_{0t}\bleftBrackets\bftilx(t) \mid \bftilx_0(t)\brightBrackets,
    \label{eq:restoration_pert_kernel}\\
    \tilp_{0t}\bleftBrackets\bftilx(t) \mid \bftilx_0(t)\brightBrackets
    &=\matN\leftBrackets \cfrac{\bfx(t)}{s(t)};\bftilx_0(t),{\sigma(t)}^2\bfI\rightBrackets,
    \label{eq:bftilx_distribution}\\
    \bftilx_0(t) 
    &= \bfx(0) + \frac{1-s(t)}{s(t)}\bfmu,
    \label{eq:bftilx0_and_x0_relation}
\end{align}
where $\matN(\bfx;\bfm,\bfSigma)$ denotes the Gaussian pdf evaluated at $\bfx$, 
with mean $\bfm$ and covariance $\bfSigma$.
We define $\bftilx(t) = \bfx(t) / s(t)$. The values of $s(t)$ and $\sigma(t)$ are as follows:
\begin{equation}
    s(t) = \exp\leftBrackets\int_{0}^t {f(\xi)\ud\xi }\rightBrackets,
    \sigma(t) = \sqrt{\int_0^t{\frac{{g(\xi)}^2}{{s(\xi)}^2}} \ud\xi}.
    \label{eq:st_and_sigmat}
\end{equation}
In our framework, $s(t)$ and $\sigma(t)$ are used instead of $f(t)$ and $g(t)$ for the design simplicity. 
By introducing the mean-adding term, \ie, $\frac{1-s(t)}{s(t)}\bfmu$, in \cref{eq:bftilx0_and_x0_relation}, the mean of $\bftxt$ approximately shifts to $\bfmu$, unlike generative DMs with a final mean of zero. Hence, the SDE in~\cref{eq:restoration_forward_sde} is named the mean-reverting SDE.
Concretely: 
\begin{itemize}
    \item At $t=0$, it is obvious that $s(0) = 1$ and $\sigma(0)=0$, ensuring $\bftilx(0)=\bftilx_0(0)=\bfx(0)$.
    \item At a large $t=T$, we require $\frac{1-s(T)}{s(T)}$ to be large enough to obscure 
    $\bfx(0)$, ensuring that $\bftilx(T)$ has a mean almost proportional to $\bfmu$ and a standard variance equal to $\sigma(T)$.
\end{itemize}
With the techniques above, we establish a diffusion process 
that bridges the target image $\bfx(0)$ and the cloudy image $\bfmu$ with noise $\bfn$, ensuring pixel-level fidelity in CR outputs. Notably, by omitting the mean-adding term (\ie, setting $s(t) = 1$), the EMRDM framework reduces to the generative DM in~\cite{karras2022elucidating}. Hence, our framework expands the boundary of generative DMs. 

See \cref{app:forward_process} for derivations of the forward process.

\noindent{\bf Backward Process.}
We use $s(t)$ and $\sigmat$ to derive the backward ODE. Based on \cref{eq:generating_backward_ode}, we have
\begin{equation}
    \ud \bftilx(t) = \lsb-\cfrac{\dot s(t)}{{s(t)}^2}\bfmu - \dot\sigma(t)\sigma(t) \scorefunc\rsb\ud t,
    \label{eq:backward_score_ode}
\end{equation}
where $\scorefunc=\tscore$ is the score function~\cite{hyvarinen2005estimation}, a vector field pointing to the higher density of data, with $\theta$ as its parameters. 
As $\scorefunc$ does not depend on the intractable form of $\log p_t \blb\bftxt\brb$~\cite{hyvarinen2005estimation}, it can be easily calculated. We use a denoiser function $D_{\theta}(\bfx;\sigma;c)$ to predict it, with $\bfx$ as the image input, $\sigma$ as the noise level input, and $c$ as the conditioning input. By training $\Dt$ as follows:
\begin{equation}
    \begin{aligned}
    L\bleftBrackets D_{\theta},\sigma(t)\brightBrackets=&\bbE_{\bftilx_0(t)\sim {p}_{\tdata}}\bbE_{\bfn\sim\matN\leftBrackets0,{\sigma(t)}^2\bfI\rightBrackets}\\
    &{\|D_{\theta}\bleftBrackets\bftilx_0(t)+\bfn;\sigma(t);c\brightBrackets - \bftilx_0(0) \|}_2^2,
    \label{eq:training_objective}    
    \end{aligned}
\end{equation}
with $p_{\text{data}}$ as the distribution of $\bftilx_0(t)$, we can acquire
\begin{equation}
    \scorefunc =\frac{D_{\theta}\bleftBrackets\bftilx(t);\sigma(t);c\brightBrackets + \frac{1-s(t)}{s(t)} \bfmu - \bftilx(t)}{{\sigma(t)}^2}.
    \label{eq:score_and_denoiser_relation}
\end{equation}
Though it is common to directly use a neural network as the denoiser $\Dt$, it is suboptimal for stable and effective training, as explained in~\cref{sec:network}. Hence, as shown in~\cref{fig:relationship}, we restructure $\Dt$ by training a different network $\Ftheta$ via the preconditioning technique.  
\cref{sec:network} provides details on $\Ftheta$, while the relationship between $\Ftheta$ and $\Dt$ is discussed in \cref{sec:preconditioning}.
By substituting \cref{eq:score_and_denoiser_relation} into \cref{eq:backward_score_ode}, we obtain
\begin{equation}
    \begin{aligned}
    &\ud \bftilx(t) = {\BBleftSquareBrackets-\cfrac{\dot s(t)}{{s(t)}^2}\bfmu - \cfrac{\dot\sigma(t)}{\sigma(t)}\times \Bigg.}\\
    &{\Bigg.\leftBrackets D_{\theta}\bleftBrackets\bftilx(t);\sigma(t);c\brightBrackets+\cfrac{1-s(t)}{s(t)}\bfmu-\bftilx(t)\rightBrackets\BBrightSquareBrackets}\ud t.
    \label{eq:backward_denoiser_ode}
    \end{aligned}
\end{equation}
See~\cref{app:backward_process} for the proof of \cref{eq:backward_score_ode,eq:training_objective,eq:score_and_denoiser_relation,eq:backward_denoiser_ode}.
We redesign the samplers based on this ODE, detailed in~\cref{sec:sampling}. Generally, as depicted in \cref{fig:relationship} (a), at time $T$, the samplers iteratively use $\Dt$ to estimate $\bfx(0)$. The output $\hat\bfx(0)$ is used in~\cref{eq:backward_denoiser_ode} to compute the next-step image $\bftilx(T-1)$ for $N$ steps, ultimately restoring the image.

\noindent{\bf Choices of $s(t)$ and $\sigma(t)$.} 
It is essential to ensure  
$\sigma(0)=0$ and
$\lim_{t\rightarrow 0}\frac{1-s(t)}{s(t)}=0$. 
We adopt the linear choice $\sigma(t)=t$ according to~\cite{karras2022elucidating}, and set $s(t) = \frac{1}{1+\alpha t}$, where $\alpha$ controls the mean reversion rate. 
Such settings yield a simpler SDE parameterization compared to prior MRDMs~\cite{luo2023image}.
\subsection{Multi-temporal Denoising Network}
\label{sec:network}
\network
\noindent{\bf Network Architecture.} 
For mono-temporal CR, previous improvements to the denoising network~\cite{peebles2023scalable,crowson2024scalable,bao2023all} can be directly used, as it is orthogonal to other modules. 
We choose HDiT~\cite{crowson2024scalable} for effectiveness and efficiency.
To adapt HDiT to CR tasks, we reset the input channels and remove the non-leaking augmentation~\cite{karras2020training} and classifier-free guidance~\cite{ho2021classifierfree}, as they are unsuitable for restoration. 
Following~\cite{saharia2022palette}, we concatenate the noisy cloudy image $\bftilx_0(t) + n$ with the condition $c$. The condition includes cloudy images and optional auxiliary modal images (\eg, SAR or IR images). 

To extend HDiT to multi-temporal CR tasks, we propose a new denoising network based on UTAE~\cite{garnot2021panoptic}
to denoise sequential images. 
As shown in \cref{fig:network} (a), we retain the main architecture of HDiT and create $L$ weight-sharing copies of the encoder and middle HDiT blocks 3, while keeping the decoder unchanged. 
In the bottleneck module, we introduce a temporal HDiT block (THDiT), 
allowing sequential feature maps to be condensed into one map. Attention masks are generated from THDiT and used to collapse the temporal dimension of the skip feature maps per resolution:
\begin{equation}
    o^i = \text{Concat}{\leftSquareBrackets\sum_{l=1}^{L}{\text{bilinear}(a^{g}_{l},i)\odot e^{i,g}_l}\rightSquareBrackets}_{g=1}^G,
    \label{eq:attention_mask_upsample_multiply}
\end{equation}
where $o^i$ is the output skipping feature map to the decoder at resolution level $i$, $a^{g}_{l}$ is the attention mask at head $g$ and time $l$, $e^{i,g}_{l}$ is the input feature map from the encoder at head $g$, time $l$ and resolution level $i$, $G$ is the number of heads, $\odot$ is the element-wise multiplication, and $\text{bilinear}(\cdot, i)$ indicates upsampling the map from the lowest resolution to level $i$.

\noindent{\bf Temporal HDiT Block.} 
THDiT is modified from the original HDiT block. As shown in \cref{fig:network} (b), we replace spatial attention with our proposed temporal fusion self-attention (TFSA) to merge sequential feature maps and generate attention masks. We also introduce rearrangement layers to ensure that the feature maps have the correct shape before entering different blocks. As the temporal dimension collapses after TFSA, we remove the residual connection. 

\noindent{\bf Temporal Fusion Self-Attention.} 
As shown in~\cref{fig:network} (c), 
TFSA adopts vanilla multi-head self-attention. 
Following L-TAE~\cite{garnot2020lightweight}, we define query, key and value matrices as $\bfQ \in \bbR^{1\times d_{k}}, \bfK=\bfX \bfW \in \bbR^{L\times d_k},\bfV =\bfX \in \bbR^{L\times C},$ respectively.
Here, we consider a single-head scenario and omit the batch size dimension for simplicity. The feature map $\bfX$ has a sequence length of $L$ and $C$ channels. Both $\bfQ$ and $\bfK$ have $d_k$ channels. 
We use $\bfX$ as $\bfV$, and project it to $\bfK$ with weights $\bfW \in \bbR^{C\times d_{k}}$. 
$\bfQ$ is set as a learnable parameter and initialized from a normal distribution, with a sequence length of $1$ to condense the temporal information.

\subsection{Preconditioning}
\label{sec:preconditioning}
In this section, we restructure the denoiser via the preconditioning technique to adaptively scale inputs and outputs according to noise variance $\sigmat$, focusing on multi-temporal CR, with the mono-temporal case covered by setting $L=1$. 
We use the superscript $l$ to represent the time point.

For training a network, it is advisable to maintain both inputs and outputs with unit variance~\cite{huang2023normalization,bishop1995neural}, thus stabilizing and enhancing the training process. While directly training denoiser $\Dt$ is not ideal for this purpose, we train a network $\Ftheta$ instead via the preconditioning technique to scale inputs and outputs to unit variance, following EDM~\cite{karras2022elucidating}. 
As shown in \cref{fig:relationship} (b), the relation between $D_{\theta}$ and $F_{\theta}$ is:
\begin{equation}
    \begin{aligned}
    &D_{\theta}\lb \{\bftilx^l\}_{l=1}^L;\sigma;c\rb=\text{mean}\lb\{\cskips\bftilx^l\}_{l=1}^L\rb\\
    &+\couts F_{\theta}\lb \{\cins\bftilx^l\}_{l=1}^L;\cnoises;c\rb,
    \label{eq:relation_D_and_F}
    \end{aligned}
\end{equation}
where $\sigma(t)$ is simplified to $\sigma$ and $\bftxt^l$ is simplified to $\bftilx^l$. The output shape of $F_{\theta}$ differs from the input shape, which requires a mean operator to reduce the temporal dimension of $\{\cskips\bftilx^l\}_{l=1}^L$. As our network can process sequential images, $\{\cins\bftilx^l\}_{l=1}^L$ does not need the mean operator. To ensure that inputs and targets have unit variance, we introduce four factors $\cins$, $\cskips$, $\couts$ and $\cnoises$ to scale the inputs and outputs governed by four hyperparameters: $\sigma_{\tdata}$ (the variance of target images), $\sigma_{\tmu}$ (the variance of cloudy images), $\sigma_{\tcov}$ (the covariance between target and cloudy images), and $L$ (sequence length):
\begin{align}
    \cin(\sigma)&=\frac{1}{\sqrt{\sigma_{\tdata}^2 + k^2 \sigma_{\tmu}^2+\sigma^2+ 2 k \sigma_{\tcov}}}, \label{eq:c_in} \\
    \cskip(\sigma)&= \frac{\sigma_{\tdata}^2 + k \sigma_{\tcov}}{\sigma_{\tdata}^2 + k^2 \sigma_{\tmu}^2 + \frac{\sigma^2}{L} + 2k\sigma_{\tcov}}, \label{eq:c_skip}\\
    \cout(\sigma) &= \sqrt{\frac{k^2\sigma_{\tmu}^2\sigma_{\tdata}^2 + \frac{\sigma^2}{L}\sigma_{\tdata}^2 - k^2\sigma_{\tcov}^2}{\sigma_{\tdata}^2 + k^2 \sigma_{\tmu}^2 + \frac{\sigma^2}{L} + 2k\sigma_{\tcov}}},\label{eq:c_out} \\
    \cnoise(\sigma) &= \frac{1}{4} \ln(\sigma),\label{eq:c_noise}
\end{align}
where $k$ represents $\kt$, and $\kt=\frac{1-s(t)}{s(t)}$. 
Notably, setting $\sigma_{\tmu} = \sigma_{\tcov} = 0$ reverts~\cref{eq:c_in,eq:c_out,eq:c_skip,eq:c_noise} to their original form in EDM. 
See~\cref{app:preconditioning} for derivations.

\subsection{Training and Sampling}
\label{sec:sampling}
This section details the training and sampling processes under the multi-temporal scenario, with the mono-temporal case covered by setting $L=1$. 

\noindent{\bf Training.}
The training process is detailed in~\cref{alg:train}.
We retain the training distribution of $\sigma$ in~\cite{karras2022elucidating} (line 2). Sequential images are then independently perturbed (lines 4 to 6) and denoised jointly (line 7). We further introduce a parameter $\lambdas$ to adjust the loss function at different noise levels during training (line 9):
\begin{equation}
    \bbE_{\sigma,\bfx(0),\bfn}
    \lsb{\lambda \lL \Dt\lb \{\bftilx_0^l + \bfn\}_{l=1}^L;\sigma,c\rb -  \bfx(0)\rL}_2^2\rsb,
    \label{eq:all_training_loss}
\end{equation}
where $\lambda$ and $\bftilx_0^l$ represent $\lambdas$ and $\bftilx_0^l(t)$, respectively. 
We set $\lambda(\sigma) = \frac{1}{{\cout(\sigma)}^2}$, in accordance with EDM~\cite{karras2022elucidating}.

\noindent{\bf Sampling.} 
As outlined in \cref{alg:test}, we design a stochastic sampler. 
It begins with the sequential sampling of noisy images (lines 2 to 3). 
Within the sampling loop, 
$\gamma_i$ is computed (line 5) to perturb the time $t_i$ to a higher noise level $\hatt_i$ (line 6).
Updated samples $\hatbfx_i^l$ at noise level $\hatt_i$ are obtained:
\begin{equation}
    \hatbfx_i^l = \bfx_i^l + \lb k(\hatt_i) - k(t_i)\rb \bfmu^l + \sqrt{\sigma(\hatt_i)^2 - \sigma(t_i)^2} \bfeps_i^l,
    \label{eq:enlarge_noise_and_mean}
\end{equation}
where $\bfeps_i^l$ denotes Gaussian noise. The Euler step (lines 10 to 12) based on~\cref{eq:backward_denoiser_ode} computes the next sample $\bfx_{i+1}^l$ for each $l$. The loop ends with a mean operator to collapse the temporal dimension of $\{\bfx_{N}^l\}_{l=1}^L$. The method includes following hyperparameters: $N$, $\Schurn$, $\Stmin$, $\Stmax$ and $\Snoise$, as in EDM. $N$ is the number of sample steps.
$\Schurn$, $\Stmin$ and $\Stmax$ control $\gamma_i$, while $\Snoise$ regulates the variance of $\bfeps_i^l$.
The stochastic sampler becomes deterministic when setting $\Schurn=0$.
In addition, we should set a range for $\sigma$ when sampling. In other words, $\sigma(t_{N-1}) = \sigmatmax$ and $\sigma(t_{0}) =\sigmatmin$. Both $\sigmatmax$ and $\sigmatmin$ are also hyperparameters. The intermediate $\sigma$ values are interpolated following EDM (Eq. 5 in~\cite{karras2022elucidating}).
See~\cref{app:sampling} for more details.
\trainCode
\sampleCode

\section{Performance Evaluation}
\label{sec:Experiment}
\comparisonTab
\figVisual
\subsection{Implementation Details}
\label{sec:implementation_details}
We conduct experiments on four datasets: 
CUHK-CR1~\cite{sui2024diffusion}, CUHK-CR2~\cite{sui2024diffusion} and SEN12MS-CR~\cite{ebel2020multisensor} for mono-temporal CR tasks; and Sen2\_MTC\_New~\cite{huang2022ctgan} for multi-temporal CR tasks with $L=3$.
MAE, PSNR, SSIM, SAM, and LPIPS are used as evaluation metrics.
We move more implementation details to~\cref{app:implementation}.

\subsection{Performance Comparison}
\label{sec:results}
All quantitative results are illustrated in~\cref{tab:all} using the optimal configuration for each model for a fair comparison. 
EMRDM surpasses all previous methods across all datasets and metrics, demonstrating its superiority. 
On the SEN12MS-CR dataset containing multi-spectral optical and auxiliary SAR images, EMRDM achieves significant improvements over existing methods. This validates its capability to exploit SAR's all-weather imaging characteristics and effectively process multi-spectral inputs. 
On the CUHK-CR1, CUHK-CR2, and Sen2\_MTC\_New datasets that mainly consist of RGB channels, EMRDM attains remarkable results across perceptual quality (LPIPS) and structural consistency metrics (SSIM, PSNR). 
Notably, it maintains performance superiority on the CUHK-CR1/CR2 datasets without auxiliary modalities, demonstrating robust CR capabilities with limited information. EMRDM further exhibits strong multi-temporal processing capability, as evidenced by leading metrics on the Sen2\_MTC\_New dataset. 
The visual results in~\cref{fig:visual} further prove the superior CR quality of EMRDM. In particular, when the input images are heavily cloud-covered, our model restores better textures, crucial for subsequent tasks after CR. 
\subsection{Ablation Study\& Parameter Effect}
\label{sec:ablation_study}
\abalationModuleTab
\noindent{\bf Effects of Modules.} We conduct ablation studies on key modules, as outlined in \cref{tab:abalation}, using models trained for 500 epochs with a deterministic sampler, setting $N=5$, $\sigmatd=1.0$, $\sigmatmin=0.001$ and $\sigmatmax=100$ for a fair comparison. 
The baseline (config \confhdr{A}) sets $s(t)=1$, reducing our method to generative DMs, with only noise images as inputs.
Config \confhdr{B} and \confhdr{C} incorporate cloudy and IR images, respectively. The results demonstrate their essential roles as conditioning inputs. Config \confhdr{D} verifies the effectiveness of the EMRDM framework in \cref{sec:framework} with $s(t)=\frac{1}{1+t}$ and $\sigmatmu=\sigmacov=0$. Incorporating preconditioning techniques proposed in \cref{sec:preconditioning} in config \confhdr{E}, with $\sigmatmu=1.0$, $\sigmacov=0.9$, results in improved performance.

\abalationParamTab
\noindent{\bf Effects of $\alpha$, $\sigmatmax$ and $N$.} \cref{tab:hyperparameter} presents the results while varying key parameters. Each model is trained for 500 epochs, with $\sigmatd=\sigmatmu=1.0$ and $\sigmacov=0.9$.
We use a deterministic sampler with $\sigmatmin=0.001$.
For $\alpha$, which controls the ratio of $\bfmu$ and $\bfn$ in the forward process, it yields the optimal results across all metrics when set to $3$.
For $\sigmatmax$, the results show that a moderate value (\eg, 100) produces almost all the best metrics.
For $N$, surprisingly, contrary to the expectations in generative DMs, a large $N$ yields poor results, while using only five steps delivers superior results across most metrics. This finding aligns with~\cite{yue2024resshift}.
\abalationFig

\noindent{\bf Effect of Samplers.} 
%
We examine our samplers in~\cref{fig:abalation_stochastic} using the $\alpha=3.0$ configuration in~\cref{tab:hyperparameter}, and setting $\sigmatmin=0.001$, $\sigmatmax=100$, and $N=5$. 
According to the upper row of \cref{fig:abalation_stochastic}, the stochastic sampler consistently outperforms the deterministic one in PSNR, with $\Snoise \in [1.000,1.020]$ and $\Schurn \ge 6.0$ achieving superior scores. 
However, high $\Schurn$ can negatively affect LPIPS and SSIM. While LPIPS is relatively insensitive to $\Snoise$, SSIM declines at higher $\Snoise$. We suggest using $\Snoise \approx 1.000$ and $\Schurn \approx 1.0$ for balanced metric performance.
According to the lower row of~\cref{fig:abalation_stochastic}, the optimal results are achieved across all metrics when $\Stmin \approx 0$. Generally, $\Stmax$ should be relatively large, such as 80 and 100.
\abalationSeqLengthTab
\abalationAttnFig

\noindent{\bf Effect of the Network.} 
We analyze the impact of $L$ on our network (see~\cref{tab:sequence_length}), with models trained using the $\alpha=3.0$ configuration in~\cref{tab:hyperparameter} and evaluated via a deterministic sampler ($\sigmatmin=0.001$, $\sigmatmax=100$, and $N=5$).
Increasing $L$ consistently boosts performance across all metrics, highlighting the benefits of multi-temporal inputs and our network's ability to process them. 
\cref{fig:abalation_attn} visualizes TFSA attention masks, with high attention scores for cloudless regions and low scores for cloudy ones. 
Regions occluded by clouds, characterized by low attention scores, correspondingly exhibit elevated scores in cloudless temporal counterparts. 
This validates TFSA's capacity to compensate for corrupted information by integrating information from spatially equivalent regions across the temporal dimension.

\section{Conclusion}
\label{sec:conclusion}

We propose a novel MRDM-based CR model named \textbf{EMRDM}. 
It offers a modular framework with updatable modules and an elucidated design space. 
With this advantage, we redesign core MRDM modules to boost CR performance, including restructuring the denoiser via a preconditioning technique and improving training and sampling processes. 
To achieve multi-temporal CR, a new network is devised to process sequential images in parallel. 
These improvements enable EMRDM to achieve superior results on mono-temporal and multi-temporal CR benchmarks. 

\section{Acknowledgments}
This work was supported in part by National Natural Science Foundation of China (No. 62202336, No. 62172300, No. 62372326), and the Fundamental Research Funds for the Central Universities (No. 2024-4-YB-03).
{
    \small
    \bibliographystyle{ieeenat_fullname}

}
\clearpage
\onecolumn
\setcounter{page}{1}
{
    \newpage
    \centering
    \Large
    \textbf{\thetitle}\\
    \vspace{0.5em}Supplementary Material \\
}
\appendix

\label{app:appendix}
\section{Derivation of formulas}
\label{app:derivation}
\subsection{Forward Process}
\label{app:forward_process}
The forward process (\ie, diffusion process) is defined as the SDE in \cref{eq:restoration_forward_sde}. The goal of this section is to derive the form of $\perturbkernel$, which is also called the perturbation kernel. We can rewrite the form of \cref{eq:restoration_forward_sde} into:
\begin{equation}
    \ud \bfx = -f(t)(\bfmu - \bfx)\ud t + g(t)\ud\bfomega_t,
    \label{eq:restoration_forward_sde_2}
\end{equation}
whose solution has already been solved in IR-SDE (Eq. (6) in 
\cite{luo2023image}), as
\begin{align}
    \perturbkernel &= \matN\blb \bfx(t);\bfm_t,v_{t}\bfI\brb, 
    \label{eq:ir_sde_1}\\
    \bfm_t = \bfmu + 
         \blb \bfx(0) - \bfmu \brb
    e^{-\bar{\theta}_{0:t}}&,
    v_t = \int_{0}^{t}{{g(\xi)}^2 e^{-2\bar{\theta}_{\xi:t}}}\udxi,\label{eq:ir_sde_2}
\end{align}
where $\bar{\theta}_{s:t} = \int_{s}^{t}{-f(\xi)}\udxi$. Thus,
\begin{align}
    \bfm_t &= \bfmu + \blb \bfx(0) - \bfmu \brb \exp \lb -\int_{0}^{t}{-f(\xi)}\udxi\rb =\bfmu + \blb \bfx(0) - \bfmu \brb s(t),
    \label{eq:st_sigmat_1}\\
    v(t)&= \int_{0}^{t}{{g(\xi)}^2\exp\lb -2 \int_{\xi}^{t}-f(z)\ud z\rb} \udxi
    =\int_{0}^{t}{\lsb
        g(\xi) \exp\lb \int_{\xi}^{t}{f(z)\ud z} \rb
    \rsb}^2 \udxi
    \label{eq:st_sigmat_2}\\
    &=\int_{0}^{t}{\lsb
        g(\xi) \exp\lb \int_{0}^{t}{f(z)\ud z} - \int_{0}^{\xi}{f(z)\ud z} \rb
    \rsb}^2 \udxi
    =\int_{0}^{t}{\lsb{
        \lb\cfrac{g(\xi)} {\exp\lb\int_{0}^{\xi}{f(z)\ud z} \rb}\rb
    }^2 \exp\lb 2\int_{0}^{t}f(z)\ud z \rb\rsb}  \udxi
    \label{eq:st_sigmat_3}\\
    &= \exp\lb 2\int_{0}^{t}f(z)\ud z \rb\int_{0}^{t}{
    {\lb\cfrac{g(\xi)} {s(\xi)}\rb}^2} \udxi 
    = {\st}^2\sigmat^2,
    \label{eq:st_sigmat_4}
\end{align}
where $\st$ and $\sigmat$ is detailed in \cref{eq:st_and_sigmat}. Hence, the perturbation kernel can be rewritten as:
\begin{align}
    \perturbkernel &= \matN \lb 
        \bfx(t);\bfmu + \st\lb \bfx \lb0\rb - \bfmu \rb, \st^2\sigmat^2\bfI
    \rb
    \label{eq:perturb_kernel_1}\\
    &=\st^{-d}\matN \lb 
        \cfrac{\bfx(t)}{s(t)};\termxwolbn, \sigmat^2\bfI
    \rb
    \label{eq:perturb_kernel_3}\\
    &=\st^{-d}\tilp_{0t}\lb\bftilx(t) \mid \bftilx_0(t)\rb,
    \label{eq:perturb_kernel_4}
\end{align}
where $d$ is the dimension of $\bfx$, $\bftilx(t)$ is equal to $\cfrac{\bfx(t)}{s(t)}$, and $\bftxzt$ along with $\tilp_{0t}$ is defined in \cref{eq:bftilx_distribution,eq:bftilx0_and_x0_relation}. \cref{eq:perturb_kernel_4} is the same as \cref{eq:restoration_pert_kernel}.
\subsection{Backward Process}
\label{app:backward_process}
As we have mentioned in \cref{sec:preliminary}, our forward SDE in \cref{eq:restoration_forward_sde} can be viewed as a special case of \cref{eq:generating_forward_sde} proposed by~\cite{song2021scorebased}, by defining $\bff(x,t) = f(t)(\bfx-\bfmu)$. Thus, the backward ODE can also be seen as a special case of \cref{eq:generating_backward_ode}. By substituting the relationship between $\bff(x,t)$ and $f(t)$ into~\cref{eq:generating_backward_ode}, we can acquire:
\begin{equation}
    \udbx = \lsb f(t)(\bfx-\bfmu) - \frac{1}{2}{g(t)}^2\score\rsb\udt,
    \label{eq:restoration_backward_ode_score}
\end{equation}
where we simplify $\bfx(t)$ to $\bfx$. According to \cref{eq:st_and_sigmat}, we can derive the relationship between $\st$, $\sigmat$ and $f(t)$,$g(t)$. This has already been demonstrated in the Eqs. (28) and (34) in~\cite{karras2022elucidating}, which is
\begin{equation}
    f(t) = \cfrac{\dotst}{\st}, g(t)=\st\sqrt{2\dotsigmat\sigmat},
    \label{eq:relation_st_sigmat_and_ft_gt}
\end{equation}
where $\dotst$ and $\dotsigmat$ are the derivatives of $\st$ and $\sigmat$, respectively. We can rewrite the form of \cref{eq:restoration_backward_ode_score} by substituting \cref{eq:relation_st_sigmat_and_ft_gt} into it:
\begin{equation}
    \udbx = \lsb 
    \cfrac{\dotst}{\st} (\bfx -\bfmu) - \st^2\dotsigmat\sigmat \score
    \rsb \udt.
    \label{eq:backward_ode_score_st_sigmat_1}
\end{equation}
Since we define $\bftxt = \cfrac{\bfx(t)}{\st}$. We can obtain 
\begin{equation}
    \bfx(t) = \st \bftxt.
    \label{eq:relation_bftxt_and_bfx}    
\end{equation}
We can differentiate both sides of \cref{eq:relation_bftxt_and_bfx}:
\begin{align}
    \cfrac{\ud\st}{\udt}\bftxt + \st \cfrac{\ud \bftxt}{\udt} &= \cfrac{\udbx(t)}{\udt},
    \label{eq:differentiate_bftilx_and_bfx_2} \\
    \dotst \bftxt \udt + \st \ud \bftxt &= \udbx(t).
    \label{eq:differentiate_bftilx_and_bfx_3} 
\end{align}
Substitute \cref{eq:differentiate_bftilx_and_bfx_3} and \cref{eq:relation_bftxt_and_bfx} into \cref{eq:backward_ode_score_st_sigmat_1}:
\begin{align}
    \dotst \bftxt \udt + \st \ud \bftxt &= \lsb 
    \cfrac{\dotst}{\st} (\st \bftxt -\bfmu) - \st^2\dotsigmat\sigmat \score
    \rsb \udt,
    \label{eq:backward_ode_score_st_sigmat_2}\\
    \st \ud \bftxt &= \lsb 
    -\cfrac{\dotst}{\st} \bfmu - \st^2\dotsigmat\sigmat \score
    \rsb \udt,
    \label{eq:backward_ode_score_st_sigmat_4}\\
    \ud \bftxt &= \lsb 
    -\cfrac{\dotst}{\st^2} \bfmu - \st\dotsigmat\sigmat \score
    \rsb \udt,
    \label{eq:backward_ode_score_st_sigmat_5}
\end{align}
The term $\score$ is the score function, which is predicted by the denoiser $\Dt$ mentioned in \cref{sec:framework}. However, we aim to use $\bftxt$ rather than $\bfx(t)$ as the input of $\Dt$. Hence, the relationship between $\tscore$ and $\score$ should be clarified. This is demonstrated as follows:
\begin{align}
    \tscore &= 
    \nabla_{{\bfx(t)}/{\st}}\log \lsb\st^{-d} p_t\lb \cfrac{\bfx(t)}{\st}\rb\rsb 
    \label{eq:relation_score_and_tscore_1}\\
    &= \st \nabla_{\bfx(t)}  \log \lsb p_t\blb \bfx(t)\brb\rsb.
    \label{eq:relation_score_and_tscore_2}
\end{align}
\cref{eq:relation_score_and_tscore_2} is based on $p_t\blb \bfx(t) \brb =\st^{-d}  p_t\lb \cfrac{\bfx(t)}{\st}\rb$, which can be derived the same as~\cref{eq:perturb_kernel_4}. \cref{eq:relation_score_and_tscore_2} can be substituted into \cref{eq:backward_ode_score_st_sigmat_5}:
\begin{align}
    \ud \bftxt &= \lsb 
    -\cfrac{\dotst}{\st^2} \bfmu - \dotsigmat\sigmat \tscore
    \rsb \udt,
    \label{eq:backward_ode_score_st_sigmat_6}
\end{align}
which aligns with \cref{eq:backward_score_ode}, with $\tscore=\scorefunc$. 

Next, we illuminate the relationship between $\tscore$ and the output of $\Dt$. Therefore, we can directly use the output of $\Dt$ within the sampling process. 
Generally, we hope that when $\Dt$ is trained to be ideal, the discrepancy between the predicted distribution and the target distribution of $\bftxt$ is minimized. 
This can be achieved using the score matching method~\cite{song2021scorebased,hyvarinen2005estimation}. 
Specifically, we regulate the score function calculated from the output of $\Dt$ to match the theoretical target score function. In other words, the training goal is to let $\scoreq=\tscore$, where we denote the target score function as $\scoreq$ and the target distribution of $\bftxt$ in the sampling process as $\qbftxt$. 
Since the integrals of $\qbftxt$ and $\pbftxt$ over the domain of $\bftxt$ are both equal to one, $\qbftxt=\pbftxt$ can be derived from $\scoreq=\tscore$.
The training goal can be achieved by optimizing the Fisher divergence~\cite{hyvarinen2009estimation,murphy2023probabilistic}, which is indicated by $\DF$. 
Assuming we are at diffusion step $t$, $\DF$ is given by:
\begin{equation}
    \fisher =
    \Eq\lsb\frac{1}{2}{\lL
        \tscore-\scoreq 
    \rL}^2\rsb.
    \label{eq:fisher_divergence}
\end{equation}
Thereby, we aim to demonstrate that optimizing~\cref{eq:fisher_divergence} is theoretically equivalent to optimizing our practical loss function $\Lt$ in~\cref{eq:training_objective}. Therefore, we can use \cref{eq:training_objective} instead of Fisher divergence. 
We select the training objective in \cref{eq:training_objective} to align with current generative DMs \cite{ho2020denoising,ramesh2022hierarchical,karras2022elucidating}, given that this objective has been proven effective \cite{karras2022elucidating}.  
\cite{vincent2011connection} proposes another elegant and scalable form of~\cref{eq:fisher_divergence}:
\begin{equation}
    \fisher = \Eqxtxz\lsb\frac{1}{2}{\lL
        \tscore-\scoreqc
    \rL}^2\rsb + \const,
    \label{eq:fisher_divergence_condition}
\end{equation}
where $\const$ is a constant, and $\Eqxtxz$ is the expectation of the joint distribution of $\bftxt$ and $\bftxzt$. 
Here, $\qcbftxt$ represents the conditional pdf of $\bftxt$ given $\bftxzt$. 
As we have the relationship between $\bftxt$ and $\bftxzt$, the concrete form of $\scoreqc$ can derived as:
\begin{align}
    &\scoreqc
    \label{eq:scoreqc_concrete_form_1}\\ 
    &= \nbftxt \log \matN\lb \bftxt;\bftilx_0(t),\sigmat^2\bfI\rb
    \label{eq:scoreqc_concrete_form_2}\\
    &=\nbftxt \log \lsb
        \tpid {\blb \dcovar \brb}^{-\frac{1}{2}} \exp \Npower
    \rsb
    \label{eq:scoreqc_concrete_form_3}\\
    &= \nbftxt \Npower
    \label{eq:scoreqc_concrete_form_4}\\
    &= -\cfrac{\bftxt - \bftxzt}{\sigmat^2}
    \label{eq:scoreqc_concrete_form_5},
\end{align}
which, along with $\tscore=\scorefunc$ and~\cref{eq:bftilx0_and_x0_relation}, can be substituted into~\cref{eq:fisher_divergence_condition}:
\begin{align}
    \fisher &= \Eqxtxz\lsb\frac{1}{2}{\lL
        \scorefunc + \cfrac{\bftxt - \bftxzt}{\sigmat^2}
    \rL}^2\rsb + \const
    \label{eq:fisher_divergence_concrete_form_1}\\
    &= \Eqxtxz\lsb\frac{1}{2}{\lL
        \scorefunc + \cfrac{\bftxt - \bfx(0) - \cfrac{1-s(t)}{s(t)} \bfmu}{\sigmat^2}
    \rL}^2\rsb + \const
    \label{eq:fisher_divergence_concrete_form_2}\\
    &= \frac{1}{2} \Eqxtxz\lsb\frac{1}{\sigmat^4}{\lL
        \sigmat^2\scorefunc + \bftxt - \cfrac{1-s(t)}{s(t)} \bfmu  - \bfx(0)
    \rL}^2\rsb + \const.
    \label{eq:fisher_divergence_concrete_form_3}
\end{align}
To achieve the alignment between the optimization results of~\cref{eq:fisher_divergence_concrete_form_3} and the training objective in~\cref{eq:training_objective}, we can unify the forms of the two objectives. 
Concretely, if we let
\begin{equation}
    \sigmat^2\scorefunc + \bftxt - \cfrac{1-s(t)}{s(t)} \bfmu = \Dtfunc,
    \label{eq:scorefunc_and_denoiser_1}
\end{equation}
then we obtain:
\begin{equation}
    \scorefunc  = \frac{1}{\sigmat^2}\lb\Dtfunc + \cfrac{1-s(t)}{s(t)} \bfmu -\bftxt \rb,
    \label{eq:scorefunc_and_denoiser_2}
\end{equation}
which formally establishes the relationship between the score function $\scorefunc$ and the denoiser output $\Dtfunc$, the same as~\cref{eq:score_and_denoiser_relation}. We can substitute~\cref{eq:scorefunc_and_denoiser_1} into~\cref{eq:fisher_divergence_concrete_form_3}:
\begin{equation}
    \fisher = \frac{1}{2} \Eqxtxz\lsb\frac{1}{\sigmat^4}{\lL
        \Dtfunc  - \bfx(0)
    \rL}^2\rsb + \const.
    \label{eq:fisher_divergence_and_denoiser}
\end{equation}
Given that $\jointq = \qcbftxt\qbftxzt$, we can acquire $\Eqxtxz\lsb\cdot\rsb=\Eqtxzt\Eqtxtc\lsb\cdot\rsb$. According to~\cref{eq:bftilx0_and_x0_relation}, $\bftxzt$ depends entirely on $\bfx(0)$, $\bfmu$ and $\st$. At any fixed diffusion step $t$, $\st$ is a specific determined value. Furthermore, $\bfx(0)$ and $\bfmu$ are drawn from the data distribution. Thus, we can denote the distribution of $\bftxzt$ as $\pdata$, as indicated in~\cref{eq:training_objective}. 
As for $\qcbftxt$, according to~\cref{eq:bftilx_distribution}, $\bftxt$ equals $\bftxzt+\bfn$, where $\bfn\sim\matN\lb0,\sigmat^2\bfI\rb$. Hence, given $\bftxzt$, $\bftxt\sim\matN\lb\bftxzt,\sigmat^2\bfI\rb$. 
Based on the aforementioned analysis, we can rewrite~\cref{eq:fisher_divergence_and_denoiser} as:
\begin{align}
    \fisher
    &= \frac{1}{2} 
        \bbE_{\bftxzt\sim \pdata}\bbE_{\bftxt\sim\matN\lb\bftxzt,\sigmat^2\bfI\rb}
        \lsb\cfrac{1}{\sigmat^4}{\lL
        \Dtfunc  - \bfx(0)
    \rL}^2\rsb + \const\\
    &=\frac{1}{2} 
        \bbE_{\bftxzt\sim \pdata}\bbE_{\bfn\sim\matN\lb0,\sigmat^2\bfI\rb}
        \lsb\cfrac{1}{\sigmat^4}{\lL
        \Dtfuncn  - \bfx(0)
    \rL}^2\rsb + \const,
    \label{eq:fisher_divergence_marginal_mean}
\end{align}
which aligns with the practical training objective in~\cref{eq:training_objective}, as $\bfx(0)=\bftilx_0(0)$, differing only by the coefficients $\cfrac{1}{2}$ and $\cfrac{1}{\sigmat^2}$. 
Note that the coefficients $\cfrac{1}{2}$ and $\cfrac{1}{\sigmat^2}$ both remain fixed at any given $t$. 
Consequently, at diffusion step $t$, optimizing $\fisher$ is theoretically equivalent to optimizing $\Lt$ in~\cref{eq:training_objective}, enabling us to directly use $\Lt$ rather than $\fisher$ as the training objective. 

By substituting~\cref{eq:scorefunc_and_denoiser_2} into~\cref{eq:backward_ode_score_st_sigmat_6}, we obtain the ODE in~\cref{eq:backward_denoiser_ode}, which is practically used in our sampling process.
\subsection{Preconditioning}
\label{app:preconditioning}
In this proof, we use $t$ to represent the diffusion step and $l$ to denote the time or the time point, in order to distinguish between these two key concepts. 
Note that $l$ is an integer, while $t$ is continuous. 
Substituting~\cref{eq:relation_D_and_F} into~\cref{eq:all_training_loss} yields:
\begin{align}
    \mL 
    &=\bbEsxn\lsb{\lambdas \lL 
    \text{mean}\lb
        \lB\cskips \bftilx^l(t)\rB_{l=1}^L\rb
    +\couts F_{\theta} -  \bfx(0)
    \rL}_2^2
    \rsb, 
    \label{eq:F_in_all_training_loss_1}\\
    &=\bbEsxn
    \lsb{\lambda(\sigma) \lL \text{mean}\lb\lB\cskip(\sigma) \termx\rB_{l=1}^L\rb +\cout(\sigma)F_{\theta} -  \bfx(0)\rL}_2^2\rsb,
    \label{eq:F_in_all_training_loss_2}\\
    &=\bbE
    \lsb\underbrace{\lambdas\couts^2}_{\text{effective weight}} \lL \underbrace{F_{\theta}}_{\text{network output}} -\underbrace{\cfrac{1}{\couts}\lb  \bfx(0) - \meantermx\rb}_{\text{effective training target}}\rL_2^2\rsb,
    \label{eq:F_in_all_training_loss_3}
\end{align}
where we omit the bracketed arguments in the functional notations $s(t)$, $\sigmat$ and $\Fthetax$ for notational simplicity. The $\bbEsxn$ is simplified to $\bbE$ in~\cref{eq:F_in_all_training_loss_3}. Note that while we have different corrupted images $\bfmu^l$ across various time points, there is only a single target $\bfx(0)$. 

Adhering to the EDM framework~\cite{karras2022elucidating}, we impose a variance normalization constraint on the training inputs of $F_{\theta}(\cdot)$, enforcing unit variance preservation at each temporal point $l$:
\begin{align}
    \tvar_{\bfx(0),\bfmu^l,\bfn^l}\lsb \cins \termx\rsb &= 1,
    \label{eq:unit_input_variance_1}\\
    \cins^2\tvar_{\bfx(0),\bfmu^l,\bfn^l}\lb \termxwob\rb &= 1,
    \label{eq:unit_input_variance_2}
\end{align}
Thus,
\begin{equation}
    \cins = \sqrt{\cfrac{1}{\tvar_{\bfx(0),\bfmu^l,\bfn^l}\lb \termxwob\rb}},
    \label{eq:unit_input_variance_3}
\end{equation}
where $\bfn^l$ is independent of $\bfx(0)$ and $\termxwobn$. However, $\bfx(0)$ and $\termxwobn$ are obviously not independent. 
Hence, we can calculate the variance of $\termxwob$:
\begin{align}
    &\tvarxmun\lb\termxwob\rb\\ 
    &= \tvarxmun\lb\termxwobn\rb + \tvarn(\bfn^l)
    \label{eq:variance_termx_1}\\
    &= \tvarx\lb\bfx\lb0\rb\rb +\tvarmu\lb\termmu\rb + 2\tcovar \lb
        \bfx \lb 0 \rb ,
        \termmu
    \rb + \tvarn(\bfn^l)
    \label{eq:variance_termx_2}\\
    &= \tvarx\lb\bfx\lb0\rb\rb +
    {\lb\cfrac{1-s}{s}\rb}^2\tvarmu\lb
        \bfmu^l
    \rb + 
    2\cfrac{1-s}{s}\tcovar \lb
        \bfx \lb 0 \rb ,
        \bfmu^l
    \rb + \tvarn(\bfn^l),
    \label{eq:variance_termx_3}
\end{align}
where  $\tcovar \lb
        \bfx \lb 0 \rb ,
        \termmu
    \rb$ is the covariance of $\bfx \lb 0 \rb$ and $\termmu$. Since $\bfn^l$ is drawn from $\matN(0,\sigma^2\bfI)$, its variance $\tvarn(\bfn^l)$ is equal to $\sigma^2$. We denote $\tvarx\lb \bfx\lb0\rb\rb$ as $\sigmatd^2$. For simplicity in derivation, we assume:
\begin{assumption}
    The variance of corrupted images at different time points remains constant, \ie $\forall l \in [1, L], \tvarmu\lb \bfmu^l \rb = \sigmatmu^2$.
    \label{assumption:variance_mean}
\end{assumption}
\begin{assumption}
    The covariance between corrupted images at different time points and the target image $\bfx(0)$ remains constant, \ie $\forall l \in [1, L], \tcovar \lb
        \bfx \lb 0 \rb ,
        \bfmu^l
    \rb = \sigmacov$.
    \label{assumption:variance_mean_x0}
\end{assumption}
Under the two assumptions, we can simplify \cref{eq:variance_termx_3} into
\begin{equation}
    \tvarxmun\lb\termxwob\rb = \sigmatd^2 + \termfracs^2\sigmatmu^2 + 2\termfracs \sigmacov + \sigma^2.
    \label{eq:final_variance_termx}
\end{equation}
According to \cref{eq:unit_input_variance_3} and \cref{eq:final_variance_termx}, we can get the value of $\cins$ as
\begin{equation}
    \cins = \cfrac{1}{
        \sqrt{\sigmatd^2 + \termfracs^2\sigmatmu^2 + 2\termfracs \sigmacov + \sigma^2}.
    }
    \label{eq:cins}
\end{equation}
\cref{eq:cins} is the same as \cref{eq:c_in}, if denoting $k=\cfrac{1-s}{s}$.

Following EDM~\cite{karras2022elucidating}, we rigorously enforce unit variance normalization on the effective training target in~\cref{eq:F_in_all_training_loss_3}: 
\begin{align}
    \tvarxmun \lsb 
        \cfrac{1}{\couts }\lb  
            \bfx(0) - \meantermx
        \rb
    \rsb &=1,
    \label{eq:unit_ftarget_variance_2} 
\end{align}
which leads to 
\begin{align}
    \couts^2 &= \tvarxmun \lsb 
        \bfx(0) - \sumxmun
    \rsb,
    \label{eq:unit_ftarget_variance_4} \\
    \couts^2 &= \tvarxmun \lsb 
        \blb 
            1 - \cskips 
        \brb \bfx(0) -
        \termfracs  \cfrac{\cskips}{L} \sum_{l=1}^{L}{\bfmu^l}
        - \cfrac{\cskips}{L} \sum_{l=1}^{L}{\bfn^l}
    \rsb,
    \label{eq:unit_ftarget_variance_5} 
\end{align}
where $\bfn^l$ is independent of both $\bfx(0)$ and $\bfmu^l$, and it is also independent across different time points. Therefore, 
\begin{align}
    \couts^2 &= \tvarxmu \lsb 
        \blb 
            1 - \cskips 
        \brb \bfx(0) -
        \termfracs  \cskipsL\summu
    \rsb + {\lb\cfrac{\cskips}{L}\rb}^2 \tvarn\lb\sum_{l=1}^{L}{\bfn^l}\rb,
    \label{eq:unit_ftarget_variance_6} \\
    \couts^2 &= \tvarxmu \lsb 
        \blb 
            1 - \cskips 
        \brb \bfx(0) -
        \termfracs  \cskipsL\summu
    \rsb + {\lb\cfrac{\cskips}{L}\rb}^2 \sum_{l=1}^{L}{\lb\tvarn\bfn^l\rb},
    \label{eq:unit_ftarget_variance_7} \\
    \couts^2 &= \tvarxmu \lsb 
        \blb 
            1 - \cskips 
        \brb \bfx(0) -
        \termfracs  \cskipsL\summu
    \rsb + \cfrac{\cskips^2}{L}\sigma^2.
    \label{eq:unit_ftarget_variance_8} 
\end{align}
Note that
\begin{align}
    &\tvarxmu \lsb 
        \icksips \bfx(0) -
        \termfracs  \cfrac{\cskips}{L} \summu 
    \rsb \\
    &= \icksips^2\sigmatd^2 + \termfracs^2 {\lb\cskipsL\rb}^2 \tvarmu \lb\summu\rb\\
    &-2 \icksips \cfrac{1-s}{s} \cskipsL \tcovar\lb \lb \bfx(0)\rb, \summu \rb.
    \label{eq:unit_ftarget_variance_9}    
\end{align}
We make another assumption for further derivations, as follows:
\begin{assumption}
    The corrupted images exhibit complete mutual dependence across all time points, \ie $\tvarmu \lb\summu\rb=\tvarmu \lb L\mu^l\rb=L^2\sigmatmu^2$.
    \label{assumption:independent_mu}
\end{assumption}
While this assumption is simplistic, as corrupted images at different times are not identical, it remains a valuable approximation for our derivation. This is because images corrupted at different time points can still exhibit significant similarity. 
The ablation experiments in \cref{sec:ablation_study} further demonstrate the effectiveness of the preconditioning method based on this assumption. Using our three assumptions and \cref{eq:unit_ftarget_variance_9}, we can derive:
\begin{align}
    &\tvarxmu \lsb 
        \icksips \bfx(0) -
        \termfracs  \cfrac{\cskips}{L} \summu 
    \rsb
    \label{eq:final_uinit_ftarget_variance_1}\\
    &=  \icksips^2\sigmatd^2 +  \termfracs^2 {\lb\cskipsL\rb}^2 L^2 \sigmatmu^2 -2 \icksips \cfrac{1-s}{s} \cskipsL L \sigmacov
    \label{eq:final_uinit_ftarget_variance_2}\\
    &=  \icksips^2\sigmatd^2 +  \termfracs^2 \cskips^2 \sigmatmu^2 -2 \icksips\cskips \cfrac{1-s}{s} \sigmacov.
    \label{eq:final_uinit_ftarget_variance_3}
\end{align}
Substitute \cref{eq:final_uinit_ftarget_variance_3} into \cref{eq:unit_ftarget_variance_8}, as follows:
\begin{equation}
    \couts^2 = \icksips^2\sigmatd^2 +  \termfracs^2 \cskips^2 \sigmatmu^2 -2 \icksips\cskips \cfrac{1-s}{s} \sigmacov + \cfrac{\cskips^2}{L}\sigma^2.
    \label{eq:final_uinit_ftarget_variance}
\end{equation}
Following EDM~\cite{karras2022elucidating}, we then obtain the optimal $\cskips$ by minimizing $\couts$, so that the errors of $F_{\theta}$ can be amplified as little as possible. This is expressed as:
\begin{equation}
    \cskips = \argmin_{\cskips}\couts = \argmin_{\cskips}\couts^2,
    \label{eq:minimize_cout_for_cskip}
\end{equation}
which is obtained by selecting $\couts \ge 0$, without loss of generality. 
To solve the optimal problem in \cref{eq:minimize_cout_for_cskip}, we set the derivative \wrt $\cskips$ to zero:
\begin{align}
    0 &=\cfrac{\ud \couts^2}{\ud \cskips} 
    \label{eq:derivative_wrt_cskip_1},\\
    0&=\cfrac{\ud \lsb\icksips^2\sigmatd^2 +  \termfracs^2 \cskips^2 \sigmatmu^2 -2 \icksips\cskips \fracs \sigmacov + \cfrac{\cskips^2}{L}\sigma^2\rsb}
    {\ud \cskips} 
    \label{eq:derivative_wrt_cskip_2},\\
    0 &= \lsb
        \sigmatd^2 + \termfracs^2\sigmatmu^2  + \cfrac{\sigma^2}{L} + 2 \fracs \sigmacov
    \rsb \cskips - \lb
        \sigmatd^2 + \fracs \sigmacov
    \rb
    \label{eq:derivative_wrt_cskip_4}.
\end{align}
Thus, we can acquire the value of $\cskips$:
\begin{equation}
    \cskips = \cfrac{\sigmatd^2 + \fracs \sigmacov}{\sigmatd^2 + \termfracs^2\sigmatmu^2  + \cfrac{\sigma^2}{L} + 2 \fracs \sigmacov},
    \label{eq:cskips}
\end{equation}
which aligns with \cref{eq:c_skip} with $k=\cfrac{1-s}{s}$.

By substituting \cref{eq:cskips} into \cref{eq:final_uinit_ftarget_variance}, we can attain the value of $\couts$:
\begin{align}
    \couts = \sqrt{\cfrac{
            \termfracs^2\sigmatmu^2\sigmatd^2 + \cfrac{\sigma^2}{L} \sigmatd^2-\termfracs^2\sigmacov^2
        }{
            \sigmatd^2 + \termfracs^2\sigmatmu^2 + \cfrac{\sigma^2}{L} + 2 \fracs \sigmacov
        }
    },
    \label{eq:couts}
\end{align}
which is the same as \cref{eq:c_out} since $k=\cfrac{1-s}{s}$.

The value of $\cnoises$ is the same as that in EDM~\cite{karras2022elucidating}, which is obtained based on experiments:
\begin{equation}
    \cnoises = \cfrac{1}{4} \ln \lb \sigma\rb.
    \label{eq:cnoises}
\end{equation}
\subsection{Sampling}
\label{app:sampling}
\dsampleCode
We present a detailed pseudocode for our stochastic sampler with arbitrary $\st$ and $\sigmat$ in~\cref{alg:dtest}, which can be regarded as an extension of~\cref{alg:test}. 
In~\cref{alg:dtest}, we individually sample the initial states, \ie $\bfx_0^l$, at each time point, from line 2 to line 3. Notably, The corrupted images $\bfmu^l$ differ across different time points. In other words, $\bfmu^{l_1} \neq \bfmu^{l_2}$ if $l_1 \ne l_2$ and $l_1, l_2 \in [1, L]$. 
From line 4 to line 15, we loop $N$ times to denoise ${\lB\bfx_0^l\rB}_{l=1}^L$. Specifically, from line 5 to line 8, we compute the value of $\gamma_i$, and $\gamma_i$ is used in line 9 to increase the noise level by adjusting $t_i$ to $\hatti$. Lines 11 to 12 involve performing stochastic perturbation on $\bfx_0^l$ at each time point $l$, using~\cref{eq:enlarge_noise_and_mean}. In line 14, we use~\cref{eq:backward_denoiser_ode} to evaluate $\cfrac{\ud\bftxt}{\udt}$ at diffusion step $\hatti$ and time point $l$. 
The denoiser $\Dt$ takes images from all time points, \ie $\allimg$, as its input, since it can denoise sequential images in parallel as discussed in~\cref{sec:network}. 
By integrating information across time points, $\Dt$ achieves improved results,  aided by the TFSA module discussed in~\cref{sec:network}.
We then apply an Euler step in line 15 to calculate 
the next-step image $\bfx_{i+1}^l$. 
Finally, we use a mean operator to reduce the temporal dimension of $\allimgN$, where $\allimgN \in \bbR^{L\times C\times H\times W}$ and $\bfx_{N}\in \bbR^{C\times H\times W}$, omitting batch size for clarity. 

In~\cref{alg:dtest}, there are seven key hyperparameters: $N$, $\Stmin$, $\Stmax$, $\Snoise$, $\Schurn$, $\sigmatmin$, and $\sigmatmax$, as mentioned in~\cref{sec:sampling}. Here we add some details. The $\Stmin$ and $\Stmax$ define the range for the stochastic sampling steps. Concretely, as shown from line 5 to line 8, if $t_i$ falls outside $[\Stmin, \Stmax]$, $\gamma_i$ is set to $0$. As a result, $\hatti$ is set to $t_i$ (line 9), leading to $\hat\bfx_i^l=\bfx_i^l$, thus reducing the stochastic sampler to its deterministic counterpart. If $t_i$ is within $[\Stmin, \Stmax]$, regular stochastic sampling occurs. 
$\Schurn$, along with $N$, controls the value of $\gamma_i$ in line 6, influencing the extent of stochastic perturbation in line 12. 
This approach is improved from the stochastic sampler in EDM~\cite{karras2022elucidating} by removing the $\gamma_i$ upper limit ($\sqrt{2} -1$ in EDM). Since our method yields larger $\gamma_i$ due to small $N$, removing this limit can prevent restricting randomness. The effectiveness of this modification is demonstrated in~\cref{sec:ablation_study}.
\trainAllDetails
\section{Detailed Related Work}
\label{app:drelatedwork}
In~\cref{sec:relatedwork}, we provided a brief overview of related work. Here, we offer a more comprehensive introduction.

\subsection{Cloud Removal}
\label{app:cloudremoval}
\noindent{\bf Traditional Methods.}
Traditional CR methods, with the use of mathematical transform \cite{hu2015thin,xu2019thin}, physical principles \cite{xu2015thin,wang2019detection}, information cloning \cite{ramoino2017ten, lin2012cloud}, offer great interpretability. However, they tend to underperform in comparison to deep learning techniques, which limits their practical applications. 

\noindent{\bf GAN-based Methods.}
Current deep learning-based CR methods primarily use GANs, with cGANs~\cite{mirza2014conditional} and Pix2Pix~\cite{isola2017image-to-image} as the vanilla paradigm. In CR tasks~\cite{enomoto2017filmy,grohnfeldt2018conditional,bermudez2018sar}, both cloudy images and noise are fed into the generator to produce a cloudless image. The ground truth or predicted cloudless images, along with the cloudy image, are fed into the discriminator, which determines whether the input includes the ground truth image. Through adversarial training, the generator learns to produce nearly real cloudless images.
To improve cGANs for CR tasks, SpA GAN~\cite{pan2020cloud} introduces a Spatial Attentive Network (SPANet) that incorporates a spatial attention mechanism in its generator to improve CR performance. The Simulation-Fusion GAN~\cite{gao2020cloud} further improves CR performance by integrating SAR images. It operates in two stages: first, it employs a specific convolutional neural network (CNN) to convert SAR images into optical images; then, it fuses the simulated optical images, SAR images, and original cloudy optical images using a GAN-based framework to reconstruct the corrupted regions. 
TransGAN-CFR~\cite{li2023transformer} proposes an innovative transformer-based generator with a hierarchical encoder-decoder network. This design includes transformer blocks~\cite{vaswani2017attention} using a non-overlapping window multi-head self-attention (WMSA) mechanism and a modified feed-forward network (FFN). SAR images are also integrated with cloudy images in this network, and a new triplet loss is introduced to improve CR capabilities.

\noindent{\bf DM-based Methods.}
Diffusion Models (DMs), a new type of generative model, have outperformed GANs in image generation tasks~\cite{dhariwal2021diffusion} and shown potential in image restoration tasks~\cite{li2023diffusion}, including CR. Current diffusion-based CR methods mostly adhere to the basic DM framework.
Concretely, DDPM-CR~\cite{jing2023denoising} leverages the DDPM~\cite{ho2020denoising} architecture to integrate both cloudy optical images and SAR images to extract DDPM features. The features are then used for cloud removal in the cloud removal head. 
DiffCR~\cite{zou2024diffcr} introduces an efficient time and condition fusion block (TCFBlock) for building the denoising network and a decoupled encoder for 
extracting features from conditional images (\eg SAR images) to guide the DM generation process.
SeqDM~\cite{zhao2023cloud} is designed for multi-temporal CR tasks. It comprises a new sequential-based training and inference strategy (SeqTIS) that processes sequential images in parallel. 
It also extends vanilla DMs to multi-modal diffusion models (MmDMs) for incorporating the additional information from auxiliary modalities (\eg SAR images).

\noindent{\bf Non-Generative Methods.}
Some non-generative methods have also been proposed for CR, serving as alternatives to GAN-based and DM-based methods. 
DSen2-CR~\cite{meraner2020cloud} employs a super-resolution ResNet~\cite{lim2017enhanced,lanaras2018super} and can function as a multi-modal model as it can process optical images and SAR images together by concatenating them as inputs.
GLF-CR~\cite{xu2022glf}, another multi-modal model, introduces a global-local fusion network to use the additional SAR information. 
Specifically, it is a dual-stream network where SAR image information is hierarchically integrated into feature maps to address cloud-corrupted areas, using global fusion for relationships among local windows and local fusion to transfer SAR features.
UnCRtainTS~\cite{ebel2023uncrtaints} is designed for both multi-temporal and mono-temporal CR tasks. It includes an encoder for all time points, an attention-based temporal aggregator for fusing sequential observations, and a mono-temporal decoder. 
The model incorporates multivariate uncertainty quantification to enhance CR capabilities. The version with uncertainty quantification is called UnCRtainTS $\sigma$, as shown in~\cref{tab:all}, while the one with simple L2 loss is named UnCRtainTS L2, as shown in~\cref{fig:visual}.
\subsection{Diffusion Models}
\label{app:diffusionmodels}
\noindent{\bf Generative DMs}
DMs are initially applied to image generation.
The vanilla DM, known as DDPM, is proposed by~\cite{ho2020denoising}.
Concurrently, Song \etal propose NCSN~\cite{song2019generative}, a generative model similar to DDPM, by estimating gradients of the data distribution.
Song \etal further clarify the underlying principles of DMs using score matching methods~\cite{song2021scorebased}, unifying DDPM as the VP condition and NCSN as the VE condition.
EDM~\cite{karras2022elucidating} criticizes that the theory and practice of conventional generative DMs~\cite{song2021scorebased} are unnecessarily complex and simplify DMs by presenting a clear design space to separate the design choices of various modules, integrating both VP and VE DMs. They also redesign most key modules within their EDM to further enhance the generation abilities.
Additional improvements include faster sampling~\cite{lu2022dpm,lu2022dpmpp}, new denoising networks~\cite{peebles2023scalable,crowson2024scalable}, and adjusted training loss weights~\cite{hang2023efficient}.
Our denoising network is based on HDiT~\cite{crowson2024scalable}, which employs a scalable hourglass transformer as the denoising network, effectively generating high-quality images in the pixel space.

\noindent{\bf Restoration DMs}
Building on the success of DMs in image generation, researchers have investigated their application in image restoration~\cite{li2023diffusion}. The restoration DM can be categorized into supervised and zero-shot learning methods, as discussed in~\cref{sec:relatedwork}. 
The first type is more relevant to our work, as our method adopts the supervised learning paradigm. 
Early supervised methods condition DMs on low-quality reference images by 
simply concatenating them with noise as the input to the denoising network, as demonstrated in SR3~\cite{saharia2022image} and Palette~\cite{saharia2022palette}. 
Later improvements focus on conditioning the models on pre-processed reference images and features, as seen in CDPMSR~\cite{niu2023cdpmsr} and IDM~\cite{gao2023implicit}.
A significant advancement comes from methods that modify the diffusion process itself to incorporate conditions.
Specifically, IR-SDE~\cite{luo2023image} introduces a mean-reverting SDE to define the forward process and derives the corresponding backward SDE, enabling generation from noisy corrupted images rather than pure noise and leading to improved restoration results. Refusion~\cite{luo2023refusion} enhances this approach by optimizing network architecture, incorporating VAE~\cite{kingma2013auto} for image compression, \etc.
ResShift~\cite{yue2024resshift} and RDDM~\cite{liu2024residual} both adopt the DDPM framework (\ie the VP condition). Similar to IR-SDE, they modify the forward process to incorporate both noise and residuals, facilitating diffusion from target images to noisy corrupted images. Notably, within the backward process, ResShift uses a single denoising network, while RDDM employs separate networks to predict noise and residuals.
Similar strategies have also been employed by InDI~\cite{delbracio2023inversion}, I2SB~\cite{liu2023I2SB}, \etc.
\cuhkAppFig
\cuhkVTAppFig
\SENMSCRAppFig
\SenMTCNewAppFig
\samplersFig
\hyperparamsFig
\section{Experiments}
\label{app:experiments}
\subsection{Implementation Details}
\label{app:implementation}
\subsubsection{Datasets}
\label{app:datasets}
The CUHK-CR1 and CUHK-CR2 datasets, introduced by~\cite{sui2024diffusion}, consist of images captured by the Jilin-1 satellite with a size of $512\times512$. CUHK-CR1 contains $668$ images of thin clouds, while CUHK-CR2 includes $559$ images of thick clouds. These two datasets collectively form the CUHK-CR dataset. With an ultra-high spatial resolution of $0.5$ m, the images encompass four bands: RGB and near-infrared (NIR). Following~\cite{sui2024diffusion}, the CUHK-CR1 dataset is split into $534$ training and $134$ testing images, while CUHK-CR2 is divided into $448$ training and $111$ testing images. The images are in PNG format, with integer values in the range $[0, 255]$.

The SEN12MS-CR dataset, introduced by~\cite{ebel2020multisensor}, contains coregistered multi-spectral optical images with $13$ bands from Sentinel-2 satellite and SAR images with $2$ bands from Sentinel-1 satellite. 
Collected from 169 non-overlapping regions of interest (ROIs) across continents, each averaging approximately $52\times40$ $\text{km}^2$ in size, the scenes of ROIs are divided into $256\times256$ pixel patches, with $50\%$ spatial overlap.
We use 114,050 images for training, 7,176 images for validation, and 7,899 images for testing. 
The dataset split follows previous works~\cite{ebel2020multisensor,ebel2023uncrtaints}.

The Sen2\_MTC\_New dataset, introduced by~\cite{huang2022ctgan}, consists of coregistered RGB and IR images across approximately 50 non-overlapping tiles. Each tile includes around 70 pairs of cropped $256\times256$ pixel patches with pixel values ranging from 0 to $10,000$. Following~\cite{huang2022ctgan}, the dataset is divided into $2,380$ images for training, $350$ for validation, and $687$ for testing. 
\subsubsection{Pre-Processing}
\label{app:preprocessing}
As with common deep learning methods, images must be pre-processed before being input into our neural network. Given that datasets vary in their characteristics, we apply distinct pre-processing techniques to each one, following established practices. Below, we provide a detailed explanation. 

\noindent{\bf The CUHK-CR1 and CUHK-CR2 datasets.}
Following~\cite{sui2024diffusion}, we resize images from $512\times512$ pixels to $256\times256$ pixels. Subsequently, the pixel values are rescaled to a range of $[-1, 1]$.

\noindent{\bf The Sen2\_MTC\_New dataset.}
Following~\cite{huang2022ctgan}, the pixel values of images are initially scaled to the $[0, 1]$ range by dividing by $10,000$, then normalized using a mean of $0.5$ and a standard deviation of $0.5$. For the training split, data augmentation includes random flips and a 90-degree rotation every four images.

\noindent{\bf The SEN12MS-CR dataset.}
Following~\cite{ebel2022sen12ms}, the pixel values of SAR and optical images are clipped to the ranges of $[-25, 0]$ and $[0, 10000]$, respectively. However, we rescale the pixel values of all images to the range of $[-1, 1]$ to achieve centrosymmetric pixel values, which is different from~\cite{ebel2022sen12ms}.

\subsubsection{Configuration}
\label{app:configuration}
The optimal configuration is detailed in~\cref{tab:training_details}. The number of input channels is the sum of channels from noisy corrupted images, auxiliary modal images, and original corrupted images, as shown in~\cref{fig:network}. The table lists these channels as (input noisy corrupted image channels + input auxiliary modal image channels + input original corrupted image channels). 
For example, in the \textit{Input Channels} row in~\cref{tab:training_details}, $28 (= 13 + 2 + 13)$ means that the noisy corrupted image has 13 channels, the auxiliary modal image has 2 channels and the original corrupted image has 13 channels. 
Notably, in CUHK-CR1 and CUHK-CR2 datasets, we reconstruct RGB and NIR channels following established methods, incorporating the NIR channel into the noisy corrupted image input rather than treated as auxiliary data. 
Consequently, the auxiliary modal image channel count for these datasets is zero. 
\subsubsection{Evaluation Metrics in Theory}
\label{app:theory_metrics} 
To comprehensively evaluate the performance, we employ multiple metrics including peak signal-to-noise ratio (PSNR), structural similarity index measure (SSIM)~\cite{wang2004image}, mean absolute error (MAE), spectral angle mapper (SAM)~\cite{kruse1993spectral}, and learned perceptual image patch similarity (LPIPS)~\cite{zhang2018unreasonable}.
The precise computational formulations of these metrics are as follows:
\begin{align}
    \PSNR(\bfy,\bfhaty) &= 20 \log_{10}\lb
        \cfrac{1}{\RMSE(\bfy,\bfhaty)}
    \rb,
    \label{eq:psnr}\\
    \SSIM(\bfy,\bfhaty) &= \cfrac{
        \lb 2\mu_{\bfy} \mu_{\bfhaty} + c_1\rb
        \lb 2\sigma_{\bfy\bfhaty} + c_2\rb
    }{
        \lb  \mu_{\bfy}^2+\mu_{\bfhaty}^2+c_1 \rb
        \lb  \sigma_{\bfy}^2+\sigma_{\bfhaty}^2+c_2 \rb
    },
    \label{eq:ssim}\\
    \MAE(\bfy,\bfhaty) &={\cfrac{1}{C\cdot H\cdot W}\sum_{c=1}^{C}{
        \sum_{h=1}^{H}{\sum_{w=1}^{W}{
           \lvert {\bfy_{c,h,w} - \bfhaty_{c,h,w}} \rvert
        }}
    }},
    \label{eq:mae}\\
    \SAM(\bfy,\bfhaty) &= \cos^{-1} \lb
        \cfrac{
           \sum_{c=1}^{C}{
                \sum_{h=1}^{H}{\sum_{w=1}^{W}{
                   {\bfy_{c,h,w} \cdot \bfhaty_{c,h,w}} 
                }}
            } 
        }{
            \sqrt{
            \sum_{c=1}^{C}{
                \sum_{h=1}^{H}{\sum_{w=1}^{W}{
                   \bfy_{c,h,w}^2
                }}
            } 
            \cdot
            \sum_{c=1}^{C}{
                \sum_{h=1}^{H}{\sum_{w=1}^{W}{
                   \bfhaty_{c,h,w}^2
                }}
            } 
            }
        }
    \rb,
    \label{eq:sam}\\
    \LPIPS(\bfy,\bfhaty)&=\sum_{i}\cfrac{1}{H_i\cdot W_i}\sum_{h=1}^{H}{\sum_{w=1}^{W} {\lL w_i \odot \lb \bfhaty_{h,w}^i-\bfy_{h,w}^i\rb \rL}_2^2}
    \label{eq:lpips}
\end{align}
where
\begin{equation}
    \RMSE(\bfy,\bfhaty) =\sqrt{\cfrac{1}{C\cdot H\cdot W}\sum_{c=1}^{C}{
        \sum_{h=1}^{H}{\sum_{w=1}^{W}{
           {\lb \bfy_{c,h,w} - \bfhaty_{c,h,w}\rb}^2
        }}
    }}.
    \label{eq:RMSE}
\end{equation}
Here, we denote the predicted image as $\bfhaty$ and the ground truth image as $\bfy$. with channel number $C$, height $H$ and width $W$. 
The notation $\bfy_{c,h,w}$ and $\bfhaty_{c,h,w}$ refers to a specific pixel in $\bfy$ and $\bfhaty$, indicated by subscript $c$, $h$, $w$. 
In~\cref{eq:ssim}, $\mu_{\bfy}$ and $\mu_{\bfhaty}$ represent the means, and $\sigma_{\bfy}$ and $\sigma_{\bfhaty}$ are the standard deviations of $\bfy$ and $\bfhaty$, respectively. The covariance is symbolized by $\sigma_{\bfy\bfhaty}$. The constants $c_1$ and $c_2$ stabilize the calculations. 
To compute LPIPS~\cite{zhang2018unreasonable}, a pre-trained network $\mF$ processes $\bfy$ and $\bfhaty$ to derive intermediate embeddings across multiple layers. The activations are normalized, scaled by a vector $w$, and the L2 distance between embeddings of $\bfy$ and $\bfhaty$  is calculated and averaged over spatial dimensions and layers as the final LPIPS value, as shown in~\cref{eq:lpips}.
In~\cref{eq:lpips}, $i$ indicates the layer of $\mF$, with $H_i$, $W_i$, and $w_i$ being the height, width, and scaling factor at the layer $i$. The embeddings at the position $(h, w)$ and the layer $i$ are denoted as $\bfhaty_{h,w}^i$ and $\bfy_{h,w}^i$. 
We use the official implementations of~\cite{zhang2018unreasonable} to calculate the value of LPIPS.

\subsubsection{Evaluation Metrics in Practice}
\label{app:practice_metrics}
Although the theoretical methods for these evaluation metrics are consistent across datasets, practical calculations may vary due to pre-processing, post-processing, \etc. To ensure a fair comparison, we apply different computing methods for each dataset, in line with prior research. Detailed explanations for each dataset are provided here.

\noindent{\bf The CUHK-CR1 and CUHK-CR2 datasets.} 
Following~\cite{sui2024diffusion}, we scale the pixel values of the restored and ground truth images, \ie $\bfhaty$ and $\bfy$, to the range $[0,255]$, and clamp any out-of-range values. These pixel values are then converted to unsigned integers. PSNR is calculated using all channels, while SSIM and LPIPS are first calculated for each channel and then averaged. To calculate LPIPS, we employ a pre-trained AlexNet~\cite{krizhevsky2012imagenet} as $\mF$. 

\noindent{\bf The Sen2\_MTC\_New dataset.} 
We adopt the DiffCR~\cite{zou2024diffcr} approach by rescaling the pixel values of the restored and ground truth images to the range $[0, 1000]$, clipping values outside $[0, 2000]$, and then rescaling back to $[0, 1]$. These processed images are used to compute PSNR and SSIM across all channels. 
For LPIPS, the input images are further rescaled to $[-1, 1]$ and processed using a pre-trained AlexNet~\cite{krizhevsky2012imagenet} as $\mF$.

\noindent{\bf The SEN12MS-CR dataset.}
All the images are rescaled to $[0, 1]$. Then, the rescaled images are used to compute $\PSNR$, $\SSIM$, $\MAE$, and $\SAM$, with all channels used.

\subsubsection{Reproducing Details}
\label{app:reproduce}
For closed-source methods, we use the metric values they report. In contrast, for certain open-source methods, we implement the algorithms ourselves and present visual results in~\cref{fig:visual}. When implementing previous methods, if pre-trained weights are available, we directly use them; otherwise, we retrain the models from scratch. Below, we briefly outline the implementation details of the reproduced methods.

\noindent{\bf The CUHK-CR1 and CUHK-CR2 datasets.} 
The CUHK-CR1 and CUHK-CR2 datasets are relatively new, with limited prior research~\cite{sui2024diffusion}. The authors evaluate five existing methods: SpA-GAN~\cite{pan2020cloud}, AMGAN-CR~\cite{xu2022attention}, CVAE~\cite{ding2022uncertainty}, MemoryNet~\cite{zhang2023memory}, and MSDA-CR~\cite{yu2022cloud}, alongside their proposed methods, DE-MemoryNet and DE-MSDA~\cite{sui2024diffusion}, on these two dataset. In~\cite{sui2024diffusion}, metrics for all methods are reported, with pre-trained weights provided only for MemoryNet and MSDA-CR. Consequently, we use these weights and retrain DE-MemoryNet and DE-MSDA to present visual results in~\cref{fig:visual}. DE-MSDA is excluded from~\cref{fig:visual} as it performs worse than DE-MemoryNet, despite being introduced in the same study.

\noindent{\bf The SEN12MS-CR dataset.} 
As McGAN~\cite{enomoto2017filmy} and SpA GAN~\cite{pan2020cloud} do not have pre-trained weights for this dataset, we retrain them and present the visual results in~\cref{fig:visual}. In contrast, pre-trained weights for DSen2-CR~\cite{meraner2020cloud}, GLF-CR~\cite{xu2022glf}, and UnCRtainTS~\cite{ebel2023uncrtaints} are available and have also been used for visualization in~\cref{fig:visual}. 
Notably, GLF-CR~\cite{xu2022glf} operates on $128\times128$ images, while other methods use $256\times256$ images. To ensure consistency, we divide each image into four segments, process them independently, and subsequently merge them for visualization, as shown in~\cref{fig:visual}. The performance metrics for all previous methods on this dataset are cited from~\cite{ebel2023uncrtaints} and~\cite{zou2024diffcr}.

\noindent{\bf The Sen2\_MTC\_New dataset.}
Metrics values are cited from~\cite{huang2022ctgan},~\cite{zou2023pmaa}, and~\cite{zou2024diffcr}.
We retrain McGAN~\cite{enomoto2017filmy}, Pix2Pix~\cite{isola2017image-to-image}, STGAN~\cite{sarukkai2020cloud} and UnCRtainTS~\cite{ebel2023uncrtaints}, while using pre-trained weights of CTGAN~\cite{huang2022ctgan}, PMAA~\cite{zou2023pmaa}, and DiffCR~\cite{zou2024diffcr} for visualization in~\cref{fig:visual}.

\subsection{Efficiency Analysis}
\label{app:efficiency_analysis}
\parammacTab


We first present a comparative analysis of parameter counts (\textit{Params}) and multiply-accumulate operations (\textit{MACs}) of our proposed method against recent state-of-the-art approaches. in~\cref{tab:params_and_macs} across the four datasets. 
Our analysis excludes early methods due to their significantly inferior performance compared to EMRDM and the unavailability or irreproducibility of their detailed implementations. 
All \textit{MACs} are computed with a batch size of 1 and an input image resolution of \(256 \times 256\) to ensure fair comparisons. 
It should be noted that although GLF-CR~\cite{xu2022glf} typically operates on $128\times128$ resolution images, we evaluated it at $256\times256$ resolution for efficiency analysis to maintain consistency across comparisons. 
Moreover, for DiffCR, which lacks official implementation details for the SEN12MS-CR dataset, we reproduce it on this dataset based on the description outlined in~\cite{zou2024diffcr} and report the corresponding \textit{Params} and \textit{MACs} in~\cref{tab:params_and_macs}. 
The entries labeled "DE" in~\cref{tab:params_and_macs} denote DE-MemoryNet and DE-MSDA~\cite{sui2024diffusion}, which share identical \textit{Params} and \textit{MACs}. 
The results of efficiency analysis demonstrate that EMRDM achieves performance gains with reasonable increments in \textit{Params} and \textit{MACs}, particularly for mono-temporal tasks. 
While multi-temporal tasks necessitate additional parameters of EMRDM to effectively model complex temporal dependencies in image sequences, the corresponding \textit{MACs} remain within reasonable bounds for real-world applications. 

We further analyzed the training and sampling time of EMRDM across the four datasets.
For standardization, we use the configurations in~\cref{tab:training_details} and measured training time per batch with batch size unchanged and sampling time per image with batch size changed to 1.
All experiments are conducted on a single NVIDIA RTX 4090 GPU to ensure fair comparisons.
Per-batch training times measure 1,410.5 ms (CUHK-CR1), 1,237.7 ms (CUHK-CR2), 1,230.5 ms (SEN12MS-CR), and 204.7 ms (Sen2\_MTC\_New), with per-image sampling times of 131.2 ms (CUHK-CR1), 128.0 ms (CUHK-CR2), 136.4 ms (SEN12MS-CR), and 173.1 ms (Sen2\_MTC\_New). 
These timing measurements are hardware-dependent and may fluctuate. Hence, we report only mean values. 
Notable, sampling time is particularly significant since training occurs only once, while sampling is performed repeatedly in practical CR applications. 
The measured sampling times demonstrate that EMRDM meets real-time requirements for CR applications, a critical factor for remote sensing, while delivering significant performance advantages.

\subsection{Additional Results} 
\label{app:results}
This section presents additional results, including visual examples from the CUHK-CR1, CUHK-CR2, SEN12MS-CR, and Sen2\_MTC\_New datasets in~\cref{app:fig:cuhk-cr1},~\cref{app:fig:cuhk-cr2},~\cref{app:fig:sen12mscr}, and~\cref{app:fig:senmtcnew}, respectively. 
Visual comparisons using our stochastic and deterministic samplers are shown in~\cref{app:fig:samplers}. Additionally, results under varying settings of $(\alpha, \sigmatmax, N)$ are provided in~\cref{app:fig:hyperparameters}.
{
    \newpage
    \small
    \bibliographystyle{ieeenat_fullname}

}

\end{document}